\documentclass[10pt,journal,compsoc]{IEEEtran}
\ifCLASSOPTIONcompsoc
  \usepackage[nocompress]{cite}
\else
  \usepackage{cite}
\fi
\ifCLASSINFOpdf
\else
\fi
\usepackage{epsfig}
\usepackage{amsmath}
\usepackage{amssymb}
\usepackage{multicol}
\usepackage{multirow}
\usepackage{nccmath}
\usepackage{pifont}% http://ctan.org/pkg/pifont
\usepackage{adjustbox}
\usepackage{enumitem}
\usepackage{caption}
\usepackage[table,xcdraw,dvipsnames]{xcolor}
\usepackage{booktabs}
\usepackage{ragged2e}

\usepackage{comment}
\usepackage{color}

\newcommand{\cmark}{\ding{51}}%
\newcommand{\xmark}{\ding{55}}%

\definecolor{dartmouthgreen}{rgb}{0.05, 0.5, 0.06}

\newcolumntype{x}[1]{>{\centering\arraybackslash\hspace{0pt}}p{#1}}

\makeatletter
\def\thickhline{%
  \noalign{\ifnum0=`}\fi\hrule \@height \thickarrayrulewidth \futurelet
   \reserved@a\@xthickhline}
\def\@xthickhline{\ifx\reserved@a\thickhline
               \vskip\doublerulesep
               \vskip-\thickarrayrulewidth
             \fi
      \ifnum0=`{\fi}}
\makeatother

\newlength{\thickarrayrulewidth}
\definecolor{darkgreen}{rgb}{0.0, 0.2, 0.13}
\definecolor{darkspringgreen}{rgb}{0.09, 0.45, 0.27}

\newcommand{\etal}{\textit{et al.}}
\newcommand{\etc}{\textit{etc.}}

\usepackage[breaklinks=true,bookmarks=false,linkcolor=red,colorlinks=true,citecolor=blue]{hyperref}

\begin{document}

\title{Grounded 3D-LLM with Referent Tokens}

\author{
Yilun Chen$^{*}$, \textit{Member, IEEE}, Shuai Yang$^{*}$, \textit{Student Member, IEEE}, \\ Haifeng Huang$^{*}$, \textit{Student Member, IEEE}, Tai Wang$^{}$, Runsen Xu$^{}$, \textit{Student Member, IEEE}, \\ Ruiyuan Lyu$^{}$, \textit{Student Member, IEEE}, Dahua Lin$^{}$, Jiangmiao Pang$^{\dagger}$, \textit{Member, IEEE}

\IEEEcompsocitemizethanks{
    \IEEEcompsocthanksitem Y. Chen, S. Yang, H Huang, T Wang, R. Lyu, and J. Pang are with Shanghai AI Laboratory, Shanghai, China. 
    \IEEEcompsocthanksitem R. Xu and D. Lin are the Department of Information Engineering, The Chinese University of Hong Kong, Hong Kong, China. \\
    \noindent $^{*}$ Equal Contribution. $^{\dagger}$ Corresponding Author.
}
}


\markboth{Journal of \LaTeX\ Class Files,~Vol.~14, No.~8, August~2015}%
{Shell \MakeLowercase{\textit{et al.}}: Bare Demo of IEEEtran.cls for Computer Society Journals}



\IEEEtitleabstractindextext{%
\begin{abstract}
Prior studies on 3D scene understanding have primarily developed specialized models for specific tasks or required task-specific fine-tuning. In this study, we propose \textit{Grounded 3D-LLM}, which explores the potential of 3D large multi-modal models (3D LMMs) to consolidate various 3D vision tasks within a unified generative framework. The model uses scene \textit{referent} tokens as special noun phrases to reference 3D scenes, enabling it to handle sequences that interleave 3D and textual data. Per-task instruction-following templates are employed to ensure natural and diversity in translating 3D vision tasks into language formats. To facilitate the use of \textit{referent} tokens in subsequent language modeling, we provide a large-scale, automatically curated grounded scene-text dataset with over 1 million phrase-to-region correspondences and introduce Contrastive Language-Scene Pre-training (CLASP) to perform phrase-level scene-text alignment using this data. Our comprehensive evaluation covers open-ended tasks like dense captioning and 3D question answering, alongside close-ended tasks such as object detection and language grounding. Experiments across multiple 3D benchmarks reveal the leading performance and the broad applicability of \textit{Grounded 3D-LLM}. Code and datasets are available at the \href{https://groundedscenellm.github.io/grounded_3d-llm.github.io}{project page}.
\end{abstract}

}

\maketitle

\section{Introduction}
\label{sec:introduction}
\IEEEPARstart{T}{he} pursuit of a unified model for 3D scene understanding, capable of performing various vision tasks, is a longstanding and significant challenge. Existing 3D scene-level models are typically customized for specific downstream tasks, including 3D object detection~\cite{jiang2020pointgroup, groupfree, vu2022softgroup, schult2023mask3d}, language grounding~\cite{scanrefer, 3dvg-transformer, mvt, d3net, vil3drel}, question answering~\cite{scanqa, multiclip, clipguided, 3dvlp}, and dense captioning~\cite{scan2cap, chen2023unit3d, SpaCap3D, d3net, 3djcg, vote2cap-detr, 3dvlp}. Recent 3D large multi-modal models (LMM), such as Chat-3D \cite{chat3d} and LL3DA~\cite{ll3da}, facilitate visual interactions based on detected objects for text reasoning, yet their capabilities remain confined to text-level tasks. 3D-LLM \cite{3dllm}, which extends 2D VLMs \cite{blip2} to 3D, grapples with challenges in 3D spatial reasoning due to the intrinsic knowledge derived from perspective views. This results in suboptimal performance in crucial localization tasks like object localization and language grounding, thereby limiting its practical applicability. The capacity of large multi-modal models (LMMs) to pinpoint objects or regions in complex environments is vital, especially for applications such as VR/AR, robotics, interactive embodied agents, object navigation, and manipulation. 

\begin{figure*}[t]
  \centering
  \includegraphics[width=1.\linewidth]{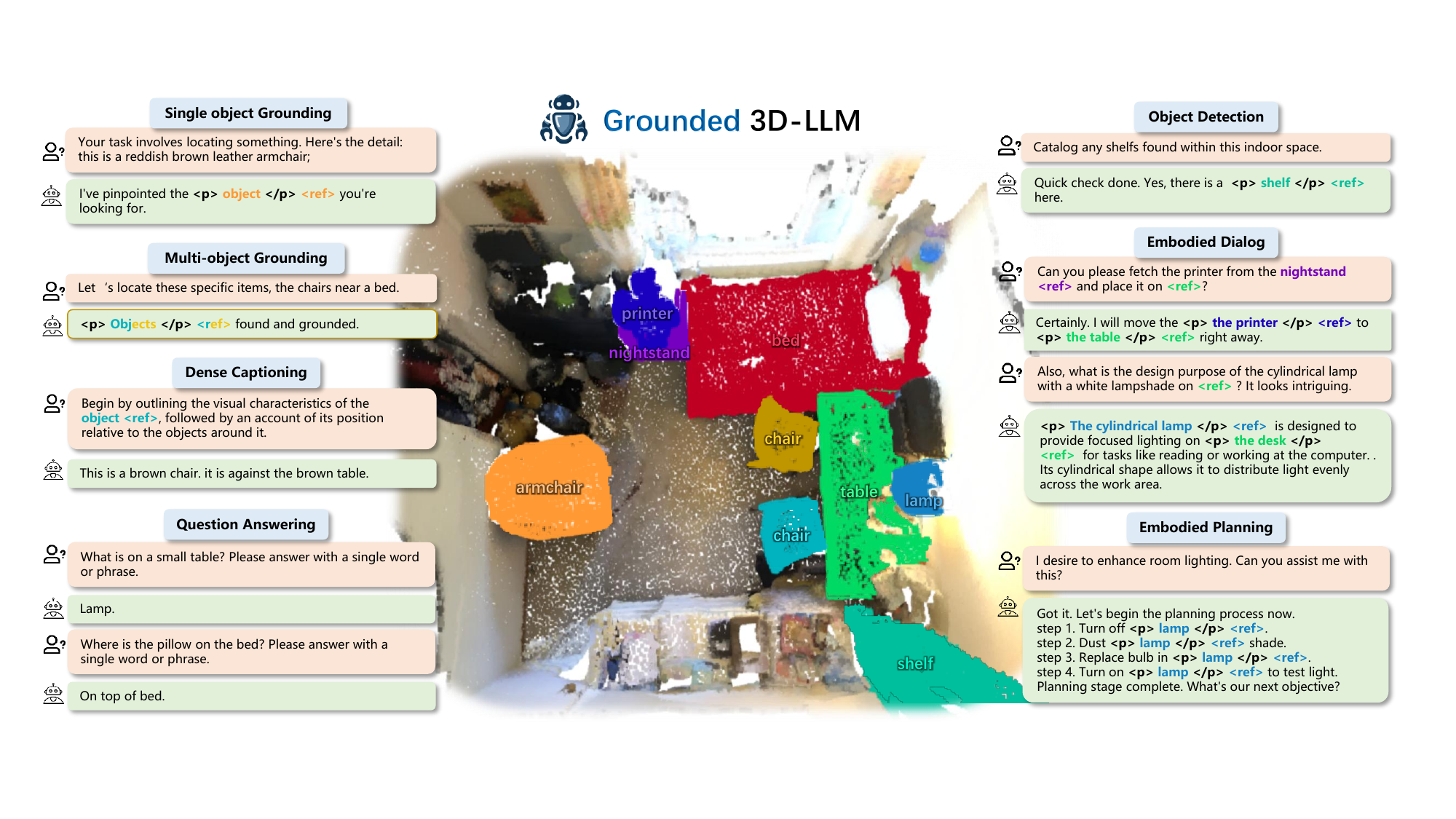}
  \caption{\textbf{Unifying scene-level 3D vision tasks via task-specific prompts in Grounded 3D-LLM framework.} The interleaved referent token \texttt{<ref>} in language modeling allows the identification of external scene referents through masked decoding in raw point clouds. Words within \texttt{<p>} and \texttt{</p>} denote the accompanying text phrases. Matching colors are shown in the figure to connect phrases with their respective object masks.}
  \label{fig: teaser}
\end{figure*}

However, the exploration of a unified model for 3D perception and reasoning remains scarce, leaving it as an open problem. In our study, by leveraging the generative modeling capability of large language models (LLM), we introduce \textbf{Grounded 3D-LLM} for integration of multiple 3D vision tasks in linguistic formats (Fig.~\ref{fig: teaser}). 

Its model design is inspired by the natural text, "Two brown chairs are near the white nightstand," which contains two noun phrases: "two brown chairs" (region) and "the white nightstand" (singular object). We refer to their corresponding 3D objects in the scene as "phraseable objects," indicating that each can be described in a text phrase and that together they are interrelated within the contextual description. Without loss of generality, we aim to represent these phraseable objects using a special token, termed the "\textit{referent} token" \texttt{<ref>}. Each referent token supports decoding to identify the corresponding objects (singular or plural) in a 3D scene. In the above example, \texttt{<ref>} can refer to items like ``the white nightstand'' or ``two brown chairs'', substituting the corresponding text tokens in the language.
The use of \textit{referent} tokens as manageable noun phrases in linguistic descriptions offers two advantages: (i) a referent token can serve as either an alternative to or a companion for text phrases without altering the contextual description, as depicted in Fig.~\ref{fig: teaser}. and (ii) it enables the localization of nouns or text phrases within the language. Thus, this design positions \textit{referents} as a unified interface, facilitating 3D grounding capabilities across various 3D vision tasks. 

To accomplish this, the language modeling of the \textit{referent} involves two high-level steps: (i) A more granular, phrase-level contrastive pre-training of the vision-language encoder in a large-scale scene-text dataset rich in phrase-level correspondence, and (ii) referent-aware instruction-following fine-tuning, which associates referent tokens with their respective object embeddings.

Specifically, unlike CLIP, which aligns text with the whole image, we first perform contrastive alignment between text primitives -- noun phrases in the contextual description -- with their corresponding visual embeddings from 3D scene encoder. This scene-level pre-training process is referred to as \textbf{C}ontrastive \textbf{LA}nguage-\textbf{S}cene \textbf{P}re-training (\textbf{CLASP}). However, existing paired scene-text data at the phrase level (referred to as \textit{grounded scene-text data} in this paper) remain scarce. To address this, we develop an automated pipeline to synthesize grounded scene-text data using a densely annotated 3D detection dataset. The resulting dataset, called Grounded Scene Caption (G-SceneCap), comprises 107K scene captions with over 1+ million phrase-level correspondence. Compared to existing scene-text datasets, it provides richer phrase-level correspondence, averaging one object every nine tokens, as shown in Table~\ref{tab: dataset statistics}. Besides, existing human-annotated 3D vision-language datasets~\cite{scanrefer, multi3drefer} are also into grounded language forms. 

After the phrase-level scene-text alignment, we conduct multi-task instruction fine-tuning for \textit{Grounded 3D-LLM} to associate the \textit{referents} with their respective phraseable objects. We convert various existing grounded scene-text datasets into instruction-following format, where referent tokens are interleaved as alternatives or companions to the text phrases (examples shown in Fig.~\ref{fig: teaser}). During training, these special \textit{referent} tokens are supervised to decode into specified object embeddings, identifying their external scene referents. These task-specific instruction-following conversions encompass a range of existing 3D vision-language tasks -- from single to multi-object grounding, instance segmentation (object detection), to 3D QA and dense captioning. We compare this model's prompt support and task versatility with previous large multi-modal models in Table~\ref{tab: 2D 3D model modality comparison}. We believe this represents a significant step toward unifying various 3D vision tasks within generative language modeling.

To summarize, our contributions are as follows:
\begin{itemize}[leftmargin=*]
    \item We propose Grounded 3D-LLM, which establishes correspondence between 3D scenes and natural language using \textit{referent} tokens. This method facilitates scene referencing and effectively models various 3D vision-language problems within a unified language modeling framework.
    % , including single- and multi-object grounding, along with introducing 3D detection for the first time.
    \item We first provide an automated, curated grounded scene caption dataset with over \textbf{1 million} phrase-level correspondences. Experiments show that CLASP, using this data in both supervised and zero-shot text settings, demonstrates broad generalization in phrase-level grounding. 
    \item Without requiring specialized models or task-specific fine-tuning, the single model Grounded 3D-LLM achieves top-tier performance in most downstream tasks among generative models, particularly in grounding problems.
\end{itemize}

\begin{table*}[t]
	\begin{center} 
        \caption{\textbf{Comparison of 3D multi-modal models.} We refer to instance segmentation as Inst.Seg., object box detection as Obj.Det., single-object grounding as Grd., point-level grounding as Point-Grd., multi-object grounding as Multi-Obj Grd., question answering as QA., and dense captioning as Cap..}\label{tab: 2D 3D model modality comparison}
         \resizebox{\linewidth}{!}{
		\begin{tabular}{lccccccccccccc}
		\toprule
            \multirow{2}{*}{Method} & \multirow{2}{*}{LLM} & \multicolumn{2}{c}{Prompts} & \multicolumn{7}{c}{Tasks} \\ \cmidrule(lr){3-4} \cmidrule(lr){5-11} 
            & & Text & Vision & Inst.Seg. & Obj.Det. & Grd. & Point-Grd. & Multi-Obj Grd. & QA. & Cap. \\ \midrule
            PointGroup~\cite{jiang2020pointgroup} & {\color{red} \xmark} & -- & --  & {\color{dartmouthgreen} \cmark} & {\color{dartmouthgreen} \cmark} & {\color{red} \xmark} & {\color{red} \xmark} & {\color{red} \xmark} & {\color{red} \xmark} & {\color{red} \xmark} \\
            Mask3D~\cite{schult2023mask3d} & {\color{red} \xmark} & -- & -- & {\color{dartmouthgreen} \cmark} & {\color{dartmouthgreen} \cmark} & {\color{red} \xmark} & {\color{red} \xmark} & {\color{red} \xmark} & {\color{red} \xmark} & {\color{red} \xmark} \\
            Multi3DRef~\cite{multi3drefer} & {\color{red} \xmark} & -- & -- & {\color{red} \xmark} & {\color{red} \xmark} & {\color{dartmouthgreen} \cmark} & {\color{red} \xmark} & {\color{dartmouthgreen} \cmark} & {\color{red} \xmark} & {\color{red} \xmark}  \\
            BUTD-DETR~\cite{butddetr} & {\color{red} \xmark} & -- & -- & {\color{red} \xmark} & {\color{dartmouthgreen} \cmark} & {\color{dartmouthgreen} \cmark} & {\color{red} \xmark} & {\color{dartmouthgreen} \cmark} & {\color{red} \xmark} & {\color{red} \xmark}  \\
            3D-VisTA~\cite{3dvista} & {\color{red} \xmark} & -- & -- & {\color{red} \xmark} & {\color{red} \xmark} & {\color{dartmouthgreen} \cmark} & {\color{red} \xmark} & {\color{red} \xmark} & {\color{dartmouthgreen} \cmark}  & {\color{dartmouthgreen} \cmark} \\ 
            \midrule
            Chat-3D~\cite{chat3d} & {\color{dartmouthgreen} \cmark} & {\color{dartmouthgreen} \cmark} & {\color{red} \xmark} & {\color{red} \xmark} & {\color{red} \xmark} & {\color{red} \xmark} & {\color{red} \xmark} & {\color{red} \xmark} & {\color{dartmouthgreen} \cmark} & {\color{dartmouthgreen} \cmark} \\
            Chat-3D v2~\cite{chat3dv2} & {\color{dartmouthgreen} \cmark} &  {\color{dartmouthgreen} \cmark} &  {\color{red} \xmark} &  {\color{red} \xmark} &  {\color{red} \xmark} & {\color{dartmouthgreen} \cmark} &  {\color{red} \xmark}&  {\color{dartmouthgreen} \cmark} &  {\color{dartmouthgreen} \cmark} & {\color{dartmouthgreen} \cmark}\\
            3D-LLM~\cite{3dllm} & {\color{dartmouthgreen} \cmark} & {\color{dartmouthgreen} \cmark} & {\color{red} \xmark} & {\color{red} \xmark} & {\color{red} \xmark} & {\color{dartmouthgreen} \cmark} & {\color{red} \xmark} & {\color{dartmouthgreen} \cmark} & {\color{dartmouthgreen} \cmark} & {\color{dartmouthgreen} \cmark} \\
            LL3DA~\cite{ll3da} & {\color{dartmouthgreen} \cmark} & {\color{dartmouthgreen} \cmark} & {\color{dartmouthgreen} \cmark} & {\color{red} \xmark} & {\color{red} \xmark} & {\color{red} \xmark} & {\color{red} \xmark} & {\color{red} \xmark} & {\color{dartmouthgreen} \cmark} & {\color{dartmouthgreen} \cmark} \\
            \midrule
            Grounded 3D-LLM & {\color{dartmouthgreen} \cmark} & {\color{dartmouthgreen} \cmark} & {\color{dartmouthgreen} \cmark} & {\color{dartmouthgreen} \cmark} & {\color{dartmouthgreen} \cmark} & {\color{dartmouthgreen} \cmark} & {\color{dartmouthgreen} \cmark} & {\color{dartmouthgreen} \cmark} & {\color{dartmouthgreen} \cmark} & {\color{dartmouthgreen} \cmark} \\
            \bottomrule
		\end{tabular}
        }
	\end{center}
\end{table*}

\section{Related Works}

\noindent\textbf{2D large multi-modal models.} 
Vision-language models~\cite{clip, blip2} are proposed to bridge visual and textual models via image-level contrastive learning. Their direct application enables open-world understanding~\cite{vild, ovdetr}. GLIP and its subsequent works~\cite{glip, detclip, groundingdino} have innovatively formulated object detection as grounding problems, leveraging detection or pseudo-labels to align semantics. 
% These models have significantly benefited from large-scale image caption and referring image-text datasets~\cite{mcoco, flickr30k, RefCOCO}. 
With stronger language models, various powerful large multi-modal models are capable of handling tasks including image captioning~\cite{kosmos2, alayrac2022flamingo}, VQA~\cite{llava, minigpt4, instructblip, cogvlm}, and object detection/segmentation~\cite{groundingdino, rasheed2023glamm}. Besides, incorporating multi-modal tokens into language modeling has been validated, including visual understanding~\cite{alayrac2022flamingo, visionllm, lai2023lisa, llavagrounding}, robot control~\cite{gato}, and interleaved text-image content generation~\cite{sun2023emu, internlmxcomposer2}. On the other hand, Chen \etal~\cite{spatialvlm} demonstrates that existing 2D VLMs lack 3D spatial reasoning, primarily due to insufficient 3D spatial data.

% Additionally, innovative approaches for handling multi-modal inputs and outputs with language have been proposed, with Flamingo~\cite{alayrac2022flamingo, openflamingo} interleaving visual and textual understanding, and other recent works~\cite{dreamllm, internlmxcomposer2} focusing on interleaved text-image content generation. 

% Some works focuses on 3D foundation models through general pre-training task~\cite{huang2023ponder, el2024probing, zhu2023ponderv2} without semantic meaning.

\noindent\textbf{3D scene understanding.}
In 3D scene understanding, prior works~\cite{huang2023ponder, zhu2023ponderv2, el2024probing} have advanced large-scale unsupervised learning for foundational models. Language is increasingly used to encapsulate user intentions due to its broad applicability, ranging from closed-set object detection~\cite{jiang2020pointgroup, groupfree, schult2023mask3d, wang2023embodiedscan, butddetr} to open-set 3D detection~\cite{chen2023clip2scene, ding2023lowis3d, pla, regionplc, zhu2023object2scene, openscene}. Typically, these 3D detectors lay the groundwork for subsequent scene-language tasks. Such tasks encompass 3D Visual Grounding~\cite{scanrefer, referit3d, mvt, multi3drefer, vil3drel, 3drp-net, 3dvg-transformer, man2024situation3d}, where models pinpoint objects within a 3D scene based on language queries. They also include 3D Question Answering~\cite{scanqa, parelli2023clip, sqa3d}, which demands robust spatial perception and reasoning. Another task, 3D Dense Captioning~\cite{scan2cap, X-trans2cap, jiao2022more}, involves localizing and describing all scene objects, necessitating a comprehensive understanding of object locations and attributes. While initial efforts were often task-specific and lacked generalizability, some methodologies~\cite{3djcg, d3net} have integrated 3D visual grounding and dense captioning tasks, leveraging their complementary nature. Recent endeavors such as 3D-VisTA~\cite{3dvista} and 3D-VLP~\cite{3dvlp} aim to establish a universal 3D visual-language framework by pre-aligning 3D scenes with language. In contrast, most approaches develop specialized models or require per-task training. They also rely on off-the-shelf 3D detectors.

\noindent\textbf{3D large multi-modal models.} 3D object-level LMMs~\cite{pointbert, xue2022ulip, liu2024openshape, uni3d} have effectively bridged the gap between 3D visuals and texts by leveraging extensive 3D object datasets~\cite{objaverse}. This has advanced the development of 3D object-level LMM~\cite{pointbind, pointllm, 3dllm, qi2023gpt4point, qi2024shapellm, uni3d, uni3dllm}. However, these models struggle with interpreting complex spatial interrelationships in 3D scenes. For scene-level LLMs, models such as Chat-3D~\cite{chat3d}, Scene-LLM~\cite{scenellm}, LL3DA~\cite{ll3da}, and LEO~\cite{embodiedgeneralist} enable scene-level dialogue, achieving notable effectiveness in tasks such as question answering and captioning. Chat-3D v2~\cite{chat3dv2} enables language grounding by using object identifiers for effective referencing. 3D-LLM~\cite{3dllm} improves scene understanding by incorporating positional embeddings and location tokens. Our approach utilizes special referent tokens for flexible scene modeling and supports diverse tasks including object detection and grounding. Additionally, our model functions as a generalist model without requiring task-specific fine-tuning.

\section{Method}
This section consists of three parts. Sec.~\ref{sec: scene-text alignment} examines vision-language alignment at three levels. Sec.~\ref{sec: Grounded 3D-LLM architecture} details the \textit{Grounded 3D-LLM}, from vision-language pre-training to multi-task instruction fine-tuning with \textit{referent} tokens. Finally, Sec.\ref{sec: grounded scene-text dataset} describes the construction of the large-scale grounded scene-text dataset for pre-training, while Sec.\ref{sec: grounded instruction-following dataset} illustrates the conversion of existing scene-text data to instruction-following formats for LLM fine-tuning.

\subsection{Preliminary: Vision-Language Correspondence} \label{sec: scene-text alignment}
The significance of vision-language alignment in 2D foundation models, exemplified by the Contrastive Language-Image Pre-training (CLIP)~\cite{clip}, is well-established. We explore three levels of vision-language alignment:
\begin{itemize}[leftmargin=*]
\item \textit{Sentence-level Correspondence:} CLIP models~\cite{clip, blip2} introduced contrastive learning between image and sentence-level text embeddings. However, the limited size ($\sim$100K-scale) of paired scene-text datasets\cite{scanrefer, referit3d, multi3drefer} hinders broader application, and the alignment lacks object-level grounding.
\item \textit{Word-level Correspondence:} Initially introduced by GLIPs~\cite{glip}, it reinterprets object detection labels as phrase grounding tasks. However, this reliance on existing detection labels does not adequately capture appearance and spatial relationships.
\item \textit{Phrase-level Correspondence:} As referred to as \textit{phrase-to-region} correspondence in prior work~\cite{mdetr}, this constraint requires linking text phrases to \textit{specific} objects (singular or plural) within a contextual description including category, attributes and \etc. However, such 3D datasets remain scarce, especially in 3D scenes.
\end{itemize}

\begin{figure*}[t]
  \centering
  \includegraphics[width=1.\linewidth]{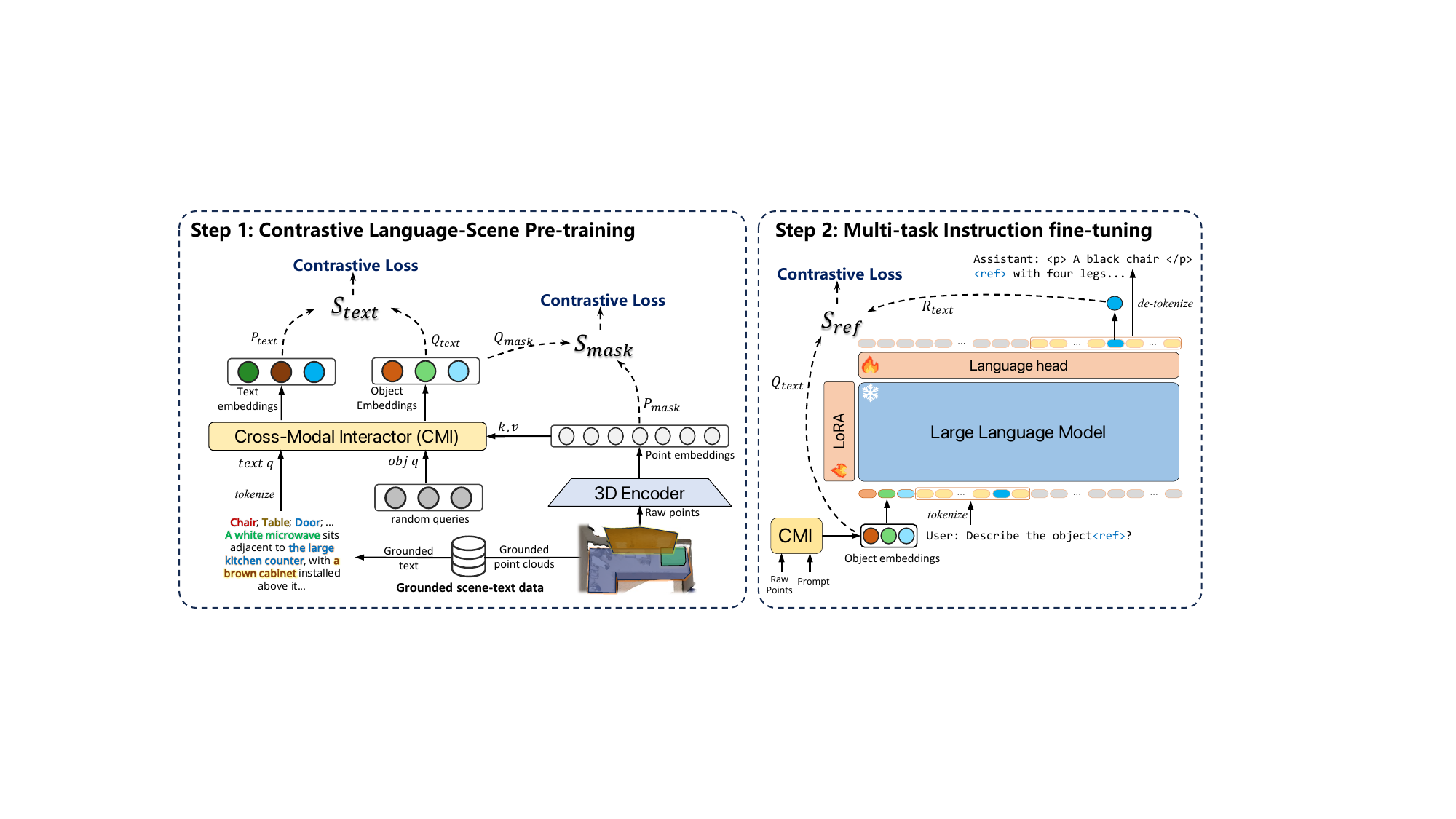}
  \caption{\textbf{Main pipeline for Grounded 3D-LLM.} The training process for Grounded 3D-LLM is divided into two key steps. Firstly, CLASP utilizes an extensive grounded scene-text dataset (at the phrase level) to pre-train a 3D point cloud encoder and a cross-modal interactor. The subsequent step involves multi-task instruction fine-tuning, which interlaces \textit{referent} tokens within the instructions and responses, thereby facilitating flexible 3D scene understanding tasks.
  }
  \label{fig: main architecture}
\end{figure*}

\noindent\textbf{Our motivation.} To enable the natural referencing of scene objects in language modeling, we aim to model phrase-level correspondences, allowing phraseable objects to be represented as referent tokens. In other words, referent tokens can substitute for or accompany noun phrases without altering the language structure. Inspired by Llava~\cite{llava}, we first adopt contrastive pre-training for the vision-language encoder at the phrase level using grounded scene-text data. During instruction-following fine-tuning, the LLM associates the newly proposed \textit{referent} token with these phraseable objects. For example, in the phrase "two brown chairs near the table," the average word embeddings of "two brown chairs" align with the 3D visual embeddings of the corresponding chairs. By inputting these text-aligned object embeddings, the LLM can reference these \textit{phraseable objects} using the special \textit{referent}. This can be formatted as \texttt{<p>}two brown chairs\texttt{</p>} \texttt{<ref>} near the table or simply as \texttt{<p>}object\texttt{</p>} \texttt{<ref>}. The token pairs \texttt{<p>} and \texttt{</p>} indicate phrase boundaries~\cite{rasheed2023glamm, llavagrounding}.

% Phrase-level correspondence offers finer scene-text alignment than sentence-level correspondence, facilitating the effective grouping of objects that share semantic attributes, including spatial relationships and appearances. Inspired by Llava~\cite{llava}, we adopts contrastive pre-training for vision-language encoder at the phrase-level using grounded scene-text data and then associate the newly proposed \textit{referent} token with these phraseable objects in language modeling. For instance, in the phrase ``two brown chairs near the table,'' the average word embeddings of ``two brown chairs'' are aligned with the 3D visual embeddings of the corresponding two chairs. Then, with the input of these text-aligned object embeddings, LLM can allows reference these \textit{phraseable objects} such as ``two brown chairs'' using the special ``referent'' token. 
% This can be represented as ``\texttt{<p>}two brown chairs\texttt{</p>} \texttt{<ref>} near the table'' or simply as ``\texttt{<p>}object\texttt{</p>} \texttt{<ref>},'' where the token pairs \texttt{<p>} and \texttt{</p>} indicate the phrases boundaries~\cite{rasheed2023glamm, llavagrounding}.

\subsection{Grounded 3D-LLM} \label{sec: Grounded 3D-LLM architecture}

Based on the aforementioned motivation, the main pipeline of \textit{Grounded 3D-LLM} is illustrated in Fig.~\ref{fig: main architecture}. To reference phraseable objects in language modeling, a vision-language model is first pre-trained on large-scale grounded scene-text data to align text phrases with their corresponding 3D objects. Subsequently, a large language model (LLM) is fine-tuned using multi-modal instruction-following data, where referent tokens serve as interleaved ``soft'' text tokens representing the phraseable objects. Per-task instruction-following templates are employed to address the diverse range of 3D vision tasks within the unified language modeling framework.

\begin{figure}[t]
  \centering
  \includegraphics[width=1.\linewidth]{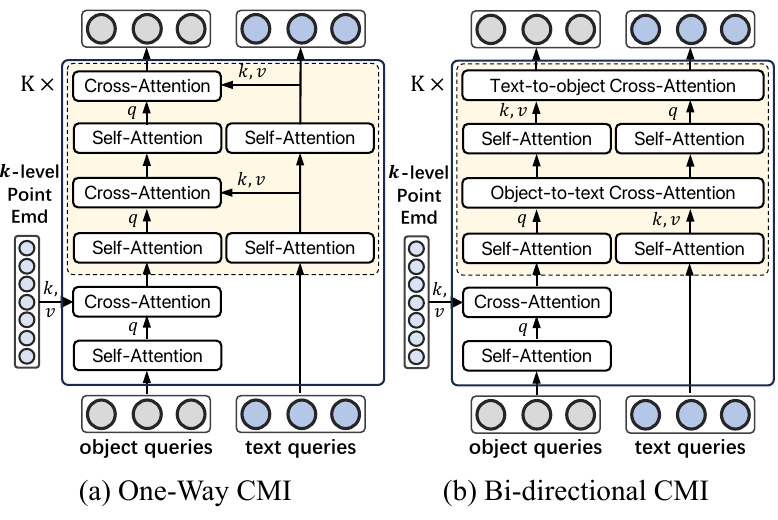}
  \caption{\textbf{Designs of Cross-Modal Interactor (CMI).} Sub-figure (b) shows the bi-directional cross-attention design inspired by GroundingDINO~\cite{groundingdino}, which is the default choice.}
  \label{fig: cmi design}
\end{figure}

\noindent\textbf{Step 1: Contrastive language-scene pre-training (CLASP).}
In the same spirit as CLIP~\cite{clip}, CLASP is pre-trained to align 3D scene embeddings with textual descriptions using \textit{grounded scene-text data}. Each paired grounded scene-text annotation contains phrase-to-region correspondences, linking noun phrases inside the text and their respective 3D object IDs (singular or plural). CLASP co-trains a 3D encoder, a text encoder, a set of object queries, and a cross-modal interactor (CMI), as shown on the left side of Fig.~\ref{fig: main architecture}. Specifically, a sparse convolutional network~\cite{spconv} serves as the 3D multi-level feature encoder, extracting point-wise embeddings $P$ as a $K$-level feature pyramid, while BERT~\cite{bert} extracts text queries $T$. For each pyramid level, a cross-modal interactor (CMI) is used to extract object embeddings conditioned on both point cloud embeddings and their respective text descriptions. Specifically, CMI interacts with $N$ learnable object queries $Q$ as proxies to connect an $M$-word text (extracted from BERT) with point clouds. Initially, these object queries interact with $k$-level point-wise embeddings via cross-attention. Then, as shown in Fig.~\ref{fig: cmi design}, to incorporate the contextual text description, they interact with text embeddings through plain one-way cross-attention or bi-directional cross-attention~\cite{groundingdino}.

\begin{figure*}[t]
  \centering
  \includegraphics[width=1.\linewidth]{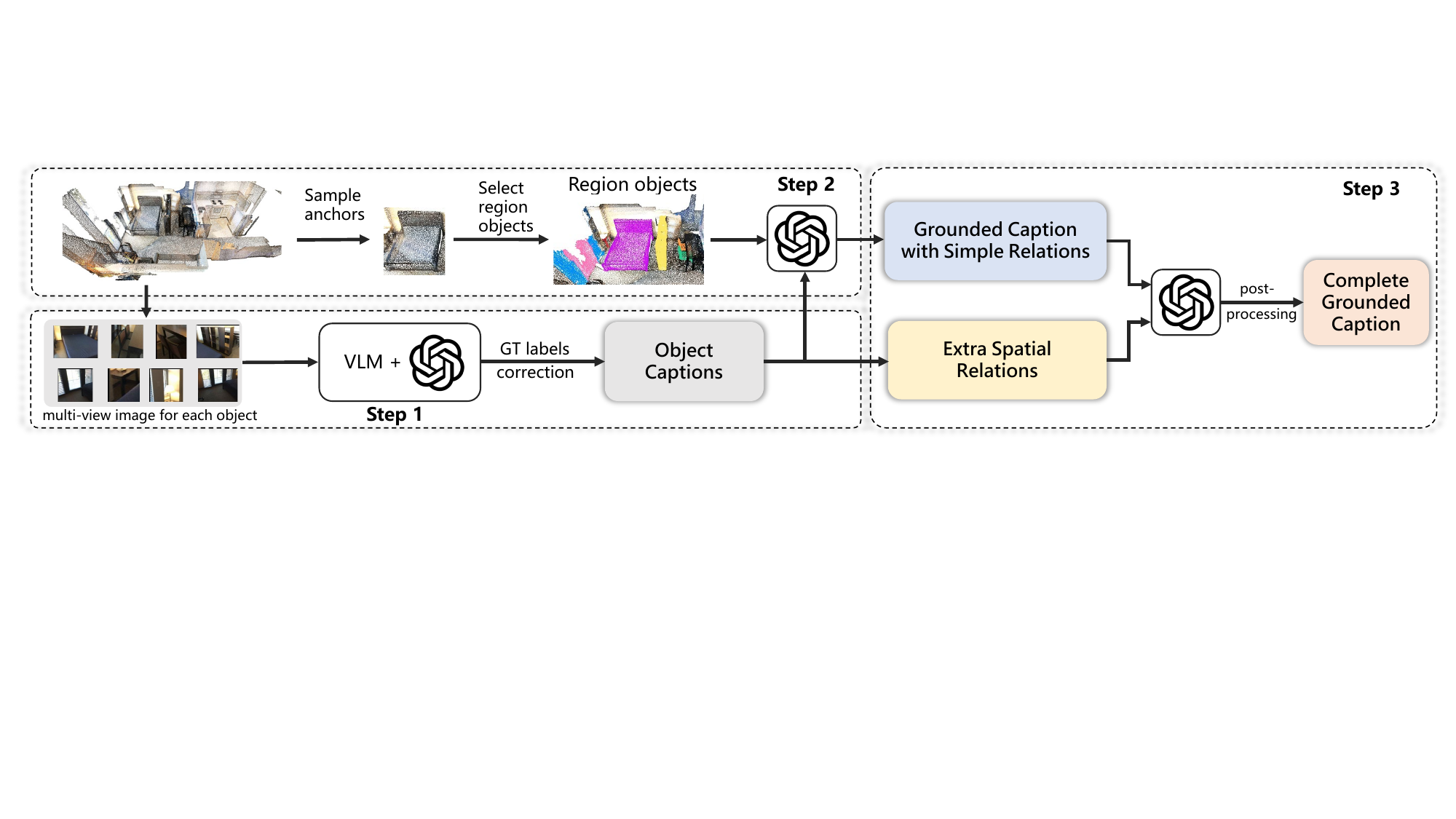}
  \caption{\textbf{Grounded Scene Caption dataset generation pipeline.} This pipeline leverages privileged knowledge in 3D, including dense object labels, multi-view images, and global spatial relationships, along with the capabilities of ChatGPT tools and spatial relationship programming tools, to establish accurate scene-level captions with grounding information.}
  \label{fig: dataset generation}
\end{figure*}

As shown in Fig.~\ref{fig: main architecture} (left), CMI yields both object and text embeddings, followed by a small head that transforms them into the textual embedding space: $P_{text}$ for positive text embeddings and $Q_{text}$ for object embeddings, as well as the mask embedding space: $P_{mask}$ and $Q_{mask}$ for mask embeddings of both, respectively:
\begin{equation}
\left[P_{mask}, Q_{mask}\right], \left[Q_{text}, P_{text}\right] = \text{CMI}(P, T, Q).
\end{equation}
As depicted in Fig.~\ref{fig: main architecture}, we compute the alignment scores $S_{mask}$ for per-point mask classification and $S_{text}$ for the textual embedding similarities between noun phrases and object queries:
\begin{equation}
S_{mask} = P_{mask} Q_{mask}^\top, \quad S_{text} = P_{text} Q_{text}^\top,
\end{equation}
where we ignore the temperature $\eta = 0.1$. The per-point mask classification for $S_{mask}$ is computed to classify each point against the object queries using \textit{bipartite matching}~\cite{detr, maskformer}. Another contrastive loss for noun phrases and object queries, $S_{text}$, is supervised by the phrase-to-region correspondence. During inference, the mapping from text phrases to point-level masks can be computed using both similarity matrices, e.g., by employing the top-scored indices as the mapping.

\noindent\textbf{Step 2: Instruction fine-tuning with \textit{referent} tokens.}
During instruction-following fine-tuning, \textit{Grounded 3D-LLM} models input and output in a unified multimodal language format, fine-tuning a LLM based on the pre-trained CLASP. Specifically, the LLM inputs scene encoding (all object query embeddings) from the frozen CLASP with a plain detection prompt~\cite{glip}. It aims to learn to use the \textit{referent} token \texttt{<ref>} to reference the scene, given a large amount of grounded instruction-following data.

Given a single grounded instruction-following data point, the text is interleaved with \textit{referent} and each referent has its mapping to some phrasesable objects (\textit{referent correspondence}). We first convert the instruction and response into the \texttt{USER}-\texttt{ASSISTANT} format before feeding them into the LLM. An example is shown below:
\begin{center}
\begin{tabular}{l}
  \texttt{[All object embeddings] USER: Describe the} \\ \texttt{object <ref> in the scene.} \\
  \texttt{ASSISTANT: <p> A black chair </p> <ref>} \\ \texttt{with four legs.}
\end{tabular}
\end{center}
To enable closed-loop dialogue with the LLM using scene referents, we model \textit{referents} as their corresponding objects in both language input and output. Specifically, for language input, the referents (visible to the model) are directly replaced with their respective object query embeddings, followed by a linear projection to match the LLM channel size. For language output, the referents serve as placeholders, with their hidden embeddings $R_{text}$ aligned with the corresponding object query embeddings. During fine-tuning, we compute the similarity matrix between the referent and all object query embeddings $Q_{text}$:
\begin{equation}
S_{ref} = R_{text}Q_{text}^\top.
\end{equation}
The contrastive loss for $S_{ref}$ is supervised with the referent correspondence and applied with the same temperature $\eta$ as CLASP. During inference, the output referents are decoded with a decoder (two MLP layers) to extract hidden embeddings $R_{text}$, retrieve the respective object queries, and obtain their associated masks through prior CLASP. 

Besides, we ensure linguistic naturalness by formatting referents with accompanying text phrases, such as ``\texttt{<p>} three nearby chairs \texttt{</p>} \texttt{<ref>}.''. If no category is provided in scene-text annotations, we use the generic term ``object'' instead. For tasks requiring grounding or localization, we append ``\texttt{(with grounding)}'' to instructions to enable referencing in the language output.

\noindent\textbf{Discussion: \textit{Referent} token as localization token.} 
Prior work, 3D-LLM~\cite{3dllm}, uses discrete location tokens, limiting localization accuracy and hindering grounded visual chat tasks. Chat-3D v2~\cite{chat3dv2} pairs object identifiers with features as text inputs but incurs high token costs and relies on off-the-shelf detectors and per-object feature extractors~\cite{uni3d}. In contrast, our model jointly extracts per-object features through large-scale, fine-grained scene-level pre-training of CLASP. The design of the Referent token allows decoding into multiple instances, reducing token costs and increasing localization efficiency. Additionally, it is the first to support point-level mask decoding.

\subsection{Generating Grounded Scene-Text Data for Pre-training} \label{sec: grounded scene-text dataset}
In this subsection, we describe two approaches to generate \textit{grounded scene-text data} for pre-training: (i) automatic generation of grounded scene captions, and (ii) transformation of existing language data into grounded formats. 

\begin{figure*}[t]
  \centering
  \includegraphics[width=1.\linewidth]{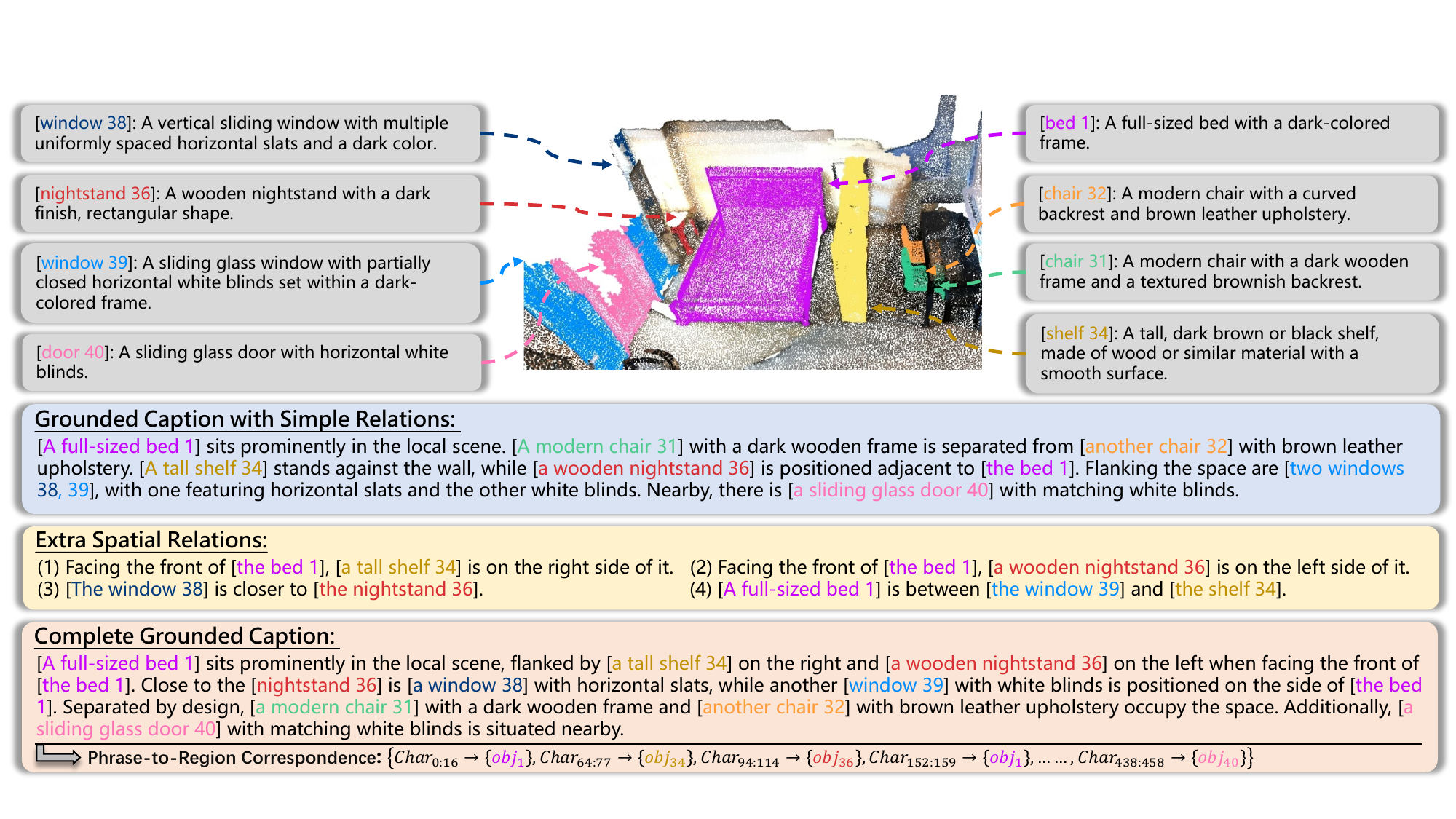}
  \caption{\textbf{Grounded Scene Caption visualization.} In each grounded text, a \textit{phrase-to-region correspondence} is formatted as ``\textsl{[object\_phrase object\_IDs]}'', where ``\textsl{object\_phrase}'' refers to a noun phrase and ``\textsl{object\_IDs}'' denotes the corresponding object IDs in the groud-truth annotation. The bottom line demonstrates explicit \textit{phrase-to-region correspondence} example between the noun phrases (character positions) and the corresponding objects. }
  \label{fig: dataset visualization}
\end{figure*}

\begin{table*}[bpt]
	\begin{center}\small
	\caption{\textbf{Grounded scene-text dataset comparison.} ``Sent.'', ``Obj.'', ``Multi.Obj.'', and ``Corr.'' refer to ``Sentence'', ``Object'', ``Multi-objects'', and ``Correspondence''. Tokens are calculated using BERT~\cite{bert}. For sentences with single-object correspondence, each text is paired with one instance.}\label{tab: dataset statistics}
         % \resizebox{1.\linewidth}{!}{
		\begin{tabular}{lcccccccc}
		\toprule
            \multirow{2}{*}{Dataset} & \multirow{2}{*}{\#3D Scan} & \multirow{2}{*}{\#Tokens} & \multirow{2}{*}{\#Text}  & \multirow{2}{*}{\shortstack{\#Tokens/Text}} & \multicolumn{3}{c}{Correspondence} \\ \cmidrule(lr){6-8}
            & & & & & \#Corr. & Type & Corr/Token \\ \midrule
            ScanRefer~\cite{scanrefer} & 0.7K & 1.17M & 52K & 22.6 & 52K & Sent.-Obj. & 4.4\% \\ 
            Sr3D~\cite{referit3d} & 1.5K & 1.10M & 84K & 13.1 & 84K & Sent.-Obj. & 7.7\% \\ 
            Sr3D+~\cite{referit3d} & 1.5K & 1.48M & 115K & 12.9 & 115K & Sent.-Obj. & 7.7\% \\ 
            Nr3D~\cite{referit3d} & 0.7K & 0.617M & 42K & 14.7 & 42K & Sent.-Obj. & 6.8\%\\ 
            % 3R-Scan~\cite{3rscan} & 1,482 & 94.3 K & Sentence & Single \\ \todo{}
            Multi3DRefer~\cite{multi3drefer} & 0.7K & 1.20M & 62K & 19.4 & 87K & Sent.-Mult.Obj. & 7.2\% \\
            ScanScribe~\cite{3dvista} & 3.0K & 18.5M & 278K & 66.5 & 278K & Sent.-Obj. & 1.5\% \\
            % SceneVerse~\cite{sceneverse} & 68K & 2.5M & Sentence & Multi & -- & -- \\ 
            \midrule
            Grounded Scene Caption & 1.5K & 9.19M & 107K & \textbf{85.9} & \textbf{1.02M} & \textbf{Phrase-Multi.Obj.} & \textbf{11.1\%} \\
            % --- Grounded ScanRefer~\cite{scanrefer} & 0.7K & 46K & Phrase & Single & 22.6 & 4.4\% \\
            % --- Grounded Multi3DRefer~\cite{multi3drefer} & 0.7K & 54K & Phrase & Multi &  & \\
            \bottomrule
		\end{tabular}
        % }
	\end{center}
\end{table*} 

\noindent\textbf{Grounded Scene Caption dataset generation.}
The dataset generation for the Grounded Scene Caption dataset (G-SceneCap) leverages ChatGPT and 2D vision-language models, utilizing dense object annotations from existing 3D scan datasets~\cite{scannet}, as illustrated in Fig.~\ref{fig: dataset generation}: 

\noindent\textit{Step 1: Bootstrapping object captions with GT label correction.} Using 3D real-scan datasets, we annotate each object with the vision-language model CogVLM~\cite{cogvlm}, using the images of the largest visible areas. Inconsistent annotations are rectified using raw instance labels. 

\noindent\textit{Step 2: Condensing objects in local scenes into a caption.} For each enumerated anchor object, we form an initial object set by randomly selecting a group of nearby objects. Their captions and coordinates \((x,y,z)\) are input into GPT-4 for captioning, which requires referencing object phrases with their object IDs in the format ``\textsl{[object\_phrase object\_ID]}'' in the caption.

\noindent\textit{Step 3: Adding rule-based relations.} To enrich scene captions, we integrate program-generated spatial relationships from Sr3D~\cite{referit3d}. By selecting an anchor object from the set in \textit{step 2}, we apply the spatial relation rules (e.g., between, supporting, nearest, back) to include related objects. GPT-4 then combines these relationships into the prior caption from \textit{step 2}.

A specific coarse-to-fine scene caption generation example with \textit{phrase-to-region correspondence}, including step-by-step intermediate outputs, is illustrated in Fig.~\ref{fig: dataset visualization}. The data generation prompts are detailed in the supplementary files.

\noindent\textbf{Grounded conversion of existing scene-text data.} 
% Prior human-annotated datasets like ScanRefer~\cite{scanrefer} and Multi3DRef~\cite{multi3drefer} identify objects using complete sentence. BUTD-DETR~\cite{butddetr} extracts positive phrases from the ScanRefer dataset using additional text models but may miss some same descriptive nouns during positive span extraction. 
In this work, we employ ChatGPT~\cite{chatgpt} to extract positive phrases for ScanRefer and Multi3DRef datasets for training and evaluation. We refer to the grounded datasets as \textit{G-ScanRefer} and \textit{G-Multi3DRefer}. The transformation of all detection labels into a fixed text prompt is called \textit{G-Detection} as GLIP~\cite{glip}.

\begin{figure*}[t]
  \centering
  \includegraphics[width=1.\linewidth]{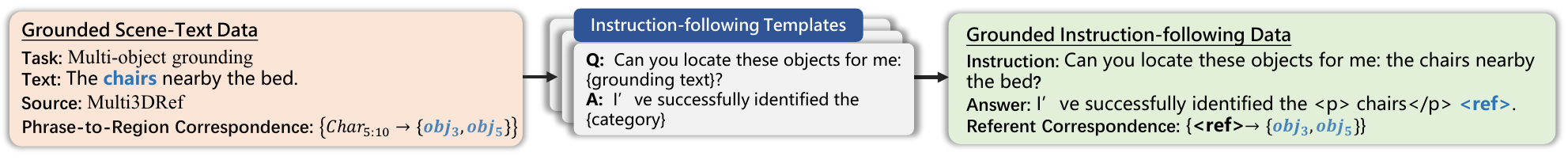}
  \caption{\textbf{Converting grounded scene-text data to instruction-following format.} This figure shows an instruction-following conversion example for the multi-object grounding task. Note that the \textit{referent correspondence} is converted from the \textit{phrase-to-region correspondence} of grounded scene-text annotation. }
  \label{fig: convert to instruction-following format}
\end{figure*}

\noindent\textbf{Comparsion with scene-text datasets.}
Llava~\cite{llava} uses GPT-4 to generate multi-modal data based on 2D boxes. Similarly, 3D-LLM~\cite{3dllm} relies only on 3D boxes to create instruction-following data. Other 3D works~\cite{pla, regionplc} add cross-view information to capture view-level differences and interactions, improving open-vocabulary detection. ScanScribe~\cite{3dvista} creates scene annotations from human scene graphs. Sceneverse~\cite{sceneverse} proposes generating scene graphs automatically, then rephrasing them into captions. Multi3DRef~\cite{multi3drefer} generates multi-object correspondence using ScanRefer annotations.

Beyond these efforts, we focus on capturing noun relationships within natural language rather than generating object-level or region-level captions. Additionally, global-view spatial relationships are considered through GPT (\textit{Step 2}) and rule-based programming (\textit{Step 3}). As a result, our generated grounded scene-text data achieves the highest object density, averaging one correspondence per nine tokens, leading to over \textit{\textbf{1.02 million} correspondences}, as shown in Table~\ref{tab: dataset statistics}. This provides rich phrase-level correspondence in natural text and allows for seamless conversion to referent correspondence in the following subsection.

\subsection{Converting to Grounded Instruction-Following Format for LLM fine-tuning} \label{sec: grounded instruction-following dataset}

In this subsection, we describe the transformation of \textit{grounded scene-text data} from Sec.~\ref{sec: grounded scene-text dataset} to the instruction-following format. 

\noindent\textbf{Converting 3D vision-language tasks to instruction-following formats.} 
For existing 3D vision-language tasks, we can transform them into instruction-following formats. An example of conversion for multi-object grounding tasks including \textit{referent correspondence} generation is shown in Fig.~\ref{fig: convert to instruction-following format}. These convertible 3D vision-language tasks include single and multi-object grounding, object detection, dense captioning, 3D QA, \etc. For each task, we utilize approximately 10-20 structured \textit{task-specific instruction-following templates}. Please refer to the supplementary file for details on the \textit{task-specific instruction-following templates}.

\noindent\textbf{Extension to embodied planning and dialogue.} To illustrate the potential applications of referent tokens in daily life and enable multi-round conversations, we prompt GPT-4 to generate embodied dialogue and planning based on the Grounded Scene Caption dataset, rather than using static templates. Unlike 3D-LLM~\cite{3dllm}, our approach produces not only plain dialogue text but also grounded instruction-following text with interleaved phrase-to-region correspondence (referent tokens). These dialogues adhere to several rules: they occur between a human and an embodied agent and involve tasks such as completing daily activities, discussing layouts, and conducting object-oriented QA based on the given caption. The embodied planning text must generate daily instructions and a step-by-step plan for interacting with objects within the scene caption. Refer to the supplementary files for the prompts used in specific dialogue data generation and the visualized results from our jointly trained model.

\begin{table*}[t]
\caption{\textbf{Evaluation of 3D scene-level LMMs as a generalist.} Entries in gray denote models specialized for specific datasets. ``Zero-shot'' denotes models used directly without fine-tuning. ``Specialist'' and ``Generalist'' categorize models fine-tuned for particular tasks versus those trained jointly. Entries in {\color[HTML]{969696}grey} denote the incorporation of ground-truth objects into the model inputs.} \label{tab: 3d llm performance}
\resizebox{\linewidth}{!}{
\begin{tabular}{lcccccccccccc}
\toprule 
\multirow{2}{*}{Method} & \multirow{2}{*}{\shortstack{Generative\\Model}} & \multicolumn{2}{c}{ScanRefer Grd.} & \multicolumn{2}{c}{Multi3DRef Grd.}  & \multicolumn{2}{c}{ScanQA} & \multicolumn{2}{c}{Scan2Cap} & \multicolumn{3}{c}{ScanNet-200 Det.} \\ \cmidrule(lr){3-4} \cmidrule(lr){5-6} \cmidrule(lr){7-8} \cmidrule(lr){9-10} \cmidrule(lr){11-13} 
 & & \scalebox{0.85}[1]{Acc@0.25} & \scalebox{0.85}[1]{Acc@0.5} & \scalebox{0.9}[1]{F1@0.25} & \scalebox{0.9}[1]{F1@0.5} & B-4 & C  & \scalebox{0.85}[1]{B-4@0.5} & \scalebox{0.85}[1]{C@0.5}  & AP & AP$_{0.25}$ & AP$_{0.5}$ \\ \midrule
{ ScanRefer~\cite{scanrefer}} & \xmark & { 37.3} & { 24.3} \\
{ MVT~\cite{mvt}} & \xmark & { 40.8} & { 33.3} \\
{ 3DVG-Trans~\cite{3dvg-transformer}} & \xmark & { 45.9} & { 34.5} \\
{ ViL3DRel~\cite{vil3drel}} & \xmark & { 47.9} & { 37.7} \\
{ M3DRef-CLIP~\cite{multi3drefer}} & \xmark & { 51.9} & { 44.7} & { 42.8} & { 38.4} \\
{ Scan2Cap~\cite{scan2cap}} & \xmark & & & & & & & { 22.4}& { 35.2} & & & \\
{ ScanQA~\cite{scanqa}} & \xmark & & & & & { 10.1}& { 64.9}  & &  &  & \\
{ 3D-VisTA~\cite{3dvista}} & \xmark & { 50.6} & { 45.8} &  &  & { 13.1} &  { 72.9} & { 34.0}& { 66.9} & & & \\ 
{ Mask3D~\cite{scanrefer}} & \xmark & & & & & & & & & { 27.4} & { 37.0} & { 42.3} \\
 \midrule \midrule
\multicolumn{5}{l}{\textit{\textbf{Zero-shot LMMs}}}\\  \addlinespace[0.1cm]
% LLaVA~\cite{llava} & -- & -- & -- & -- & 5.7 & 7.1 & 0.3 & -- & --  & -- & -- & -- \\
LLM-Grounder~\cite{llmgrounder} & \cmark & 17.1 & 5.3 & -- & -- & -- & -- & -- & --  & -- & -- & --  \\ \midrule
\multicolumn{5}{l}{\textit{\textbf{Specialist LMMs}}}\\  \addlinespace[0.1cm]
\scalebox{0.9}[1]{3D-LLM(Flamingo)}\cite{3dllm} & \cmark & 21.2 & -- & -- & -- & 7.2& 59.2 & --& -- & --  & -- & -- \\
\scalebox{0.9}[1]{3D-LLM(BLIP2-flant5)}\cite{3dllm} & \cmark & 30.3 & -- & -- & -- & 12.0& 69.4 & -- & --  & -- & --  & -- \\
Chat-3D~\cite{chat3d} & \cmark & -- & -- & -- & --& 6.4 & 53.2 & -- & -- & -- & --  & -- \\
Chat-3D v2~\cite{chat3dv2} & \cmark & 35.9 & 30.4 & -- & --  & 7.3 & 77.1 & -- & -- & -- \\
LL3DA~\cite{ll3da} & \cmark & -- & -- & -- & -- &13.5& 76.8& 36.8 & 65.2  & -- & --  & -- \\ \midrule
\multicolumn{5}{l}{\textit{\textbf{Generalist LMMs}}} \\ \addlinespace[0.1cm]
LEO~\cite{embodiedgeneralist} & \cmark & -- & -- & -- & --& {\color[HTML]{969696} 13.2} & {\color[HTML]{969696} 101.4} & \textbf{38.2} & \textbf{72.4}  & -- & -- & -- \\
LL3DA~\cite{ll3da} & \cmark & -- & -- & -- & -- & \textbf{13.3}& \textbf{75.7} & 36.0 & 62.9    & -- & -- & -- \\
% Grounded 3D-LLM & \cmark & \textbf{47.9} & \textbf{44.1} & \textbf{45.2} & \textbf{40.6}  & 35.5 & \textbf{70.6}  & \textbf{13.4}& 72.7 & \textbf{15.1} & \textbf{20.8} &  \textbf{23.5}\\ 
Grounded 3D-LLM & \cmark & \textbf{48.6} & \textbf{44.0} & \textbf{44.7} & \textbf{40.8} & \textbf{13.2}& \textbf{75.9} & 35.0 & 70.2 & \textbf{12.1} & \textbf{16.8} &  \textbf{18.9}\\ 
\bottomrule
\end{tabular}}
\end{table*}

\section{Experiments}

\subsection{Experimental Setup} \label{sec: implementation details}

\noindent\textbf{Datasets.} We evaluate our model on the ScanRefer~\cite{scanrefer} (single-object grounding), Multi3DRef~\cite{multi3drefer} (multi-object grounding), ScanNet-200 Detection~\cite{scannet200} (object detection/instance segmentation), Scan2Cap~\cite{scan2cap} (dense object captioning), and ScanQA~\cite{scanqa} (3D QA) benchmarks. These benchmarks are built on the ScanNet dataset~\cite{scannet}, a large-scale indoor 3D scene dataset with 1,513 multi-modal scenes containing 3D point clouds, image sequences, and point-level instance segmentation annotations. The dataset is split into 1,201 scenes for training and 312 for validation, and all benchmarks follow these splits.

\noindent\textbf{Metrics.} We follow prior works and adopt the commonly used metrics for these benchmarks. For ScanRefer and Multi3DRef, we compute the Intersection over Union (IoU) at thresholds of $0.25$ and $0.5$. For ScanNet-200, we report 3D instance segmentation using average precision at $25\%$, $50\%$ overlap, and mean average precision (mAP) with IoU thresholds in the $[0.5\text{:}0.95\text{:}0.05]$ range across 200 classes. For Scan2Cap and ScanQA, we evaluate language generation with \textit{BLEU}-4 and \textit{CIDEr}, weighted by IoU above $0.25$ or $0.5$.

To maintain consistency with prior evaluations across the ScanRefer, Multi3DRef, and Scan2Cap benchmarks, we extract 3D bounding boxes by taking the minimum and maximum values from the predicted masks, despite our model producing point-level predictions. For Multi3DRef, we apply a score threshold of $0.3$ to filter the predicted objects.

\begin{table}[h]
\caption{\textbf{Effects of diverse instruction-following templates on language tasks.} } \label{tab: diverse templates}
\begin{center}
% \resizebox{\linewidth}{!}{
\begin{tabular}{ccccccccccc}
\toprule 
\multirow{2}{*}{\shortstack{Diverse\\Templates}} & \multicolumn{2}{c}{ScanQA} & \multicolumn{4}{c}{Scan2Cap}  \\ \cmidrule(lr){2-3} \cmidrule(lr){4-7}
 & B-4 & C & B-4@0.25 & C@0.25 & B-4@0.5 & C@0.5 \\ \midrule 
\xmark & 11.2 & 70.8 & 31.5 & 65.8 & 29.2 & 62.1 \\
\cmark & \textbf{13.2} & \textbf{75.9} & \textbf{36.9} & \textbf{74.1} & \textbf{35.0} & \textbf{70.2} \\
\bottomrule
\end{tabular}
% }
\end{center}
\end{table}

\noindent\textbf{Implementation Details.} Unless otherwise specified, CLASP pre-training is conducted on Grounded Scene Caption, ScanNet-200 detection data, ScanRefer, and Multi3DRef data. 150 object queries via farthest point sampling (FPS) is initialized for CLASP and initialized with pre-trained 3D detector Mask3D~\cite{schult2023mask3d} to accelerate the training. We incorporate the same per-point mask classification loss $\mathcal{L}_{mask}$ with both sigmoid focal loss and Dice loss for segmentation. The overall loss function for CLASP is defined as:
\begin{equation}
    \mathcal{L}_{CLASP} = \mathcal{L}_{mask}(S_{mask}, T_{mask}) + \lambda_{cls}\mathcal{L}_{cls}(S_{text}, T_{text}),
\end{equation}
where $T_{mask}$ is computed using Hungarian matching~\cite{schult2023mask3d} and $T_{text}$ are the ground-truth mappings computed according to ground-truth \textit{phrase-to-region correspondence}. $\mathcal{L}_{cls}$ represents the sigmoid focal loss. 
% Background class because it does not correspond to a well-defined text prompt. % Further implementation details can be found in Mask3D~\cite{schult2023mask3d}. 

% \noindent\textbf{Multi-task instruction fine-tuning.} 
During instruction fine-tuning, Grounded 3D-LLM adopts Tiny-Vicuna-1B as the LLM by default, keeping it frozen except for the intermediate projection and LoRA~\cite{LoRA} layers. Its vision-language model is initialized with a pre-trained CLASP model. To minimize token costs, only 100 object queries are used. The fine-tuning leverages grounded instruction-following datasets, including ScanRefer, Multi3DRef, ScanNet-200 detection, Scan2Cap, ScanQA, scene captioning, embodied dialogue and embodied planning. An object query is considered a positive match only if its mask IoU with the ground-truth object exceeds $0.3$ otherwise the object will be ignored. The overall language loss is used as follows:
\begin{equation}
\mathcal{L}_{LLM} = \mathcal{L}_{\text{lang}} + \mathcal{L}_{ref}(S_{ref}, T_{ref}),
\end{equation}
where $L_{\text{lang}}$ is cross-entropy loss as LLAMA~\cite{touvron2023llama} and $L_{ref}$ computes the sigmoid focal loss between $S_{ref}$ and the ground-truth \textit{referent correspondence} $T_{ref}$ from the grounded annotation.

\begin{table*}[t]
\caption{\textbf{Grounding ability of Grounded 3D-LLM.} The term ``all data'' encompasses the G-ScanRefer, G-Multi3DRef, and G-SceneCap datasets used for pre-training. } \label{tab: effects of clasp for llm}
\begin{center} 
% \resizebox{0.9\linewidth}{!}{
\begin{tabular}{cccccccc}
\toprule
\multirow{2}{*}{\#} & \multirow{2}{*}{CLASP} & \multirow{2}{*}{\shortstack{Pre-training\\data}} & \multirow{2}{*}{\shortstack{\texttt{<ref>}\\type}} & \multicolumn{2}{c}{ScanRefer Grd.} & \multicolumn{2}{c}{Multi3DRef Grd.} \\ \cmidrule(lr){5-6} \cmidrule(lr){7-8} 
& & & & Acc@0.25 & Acc@0.5 & F1@0.25 & F1@0.5 \\ \midrule
(a) & \xmark & -- & One-to-Many & 6.5 & 5.8 & 6.4 & 5.3 \\
(b) & \cmark & G-ScanRefer & One-to-Many & 43.2 & 36.6 & 40.2 & 36.9 \\
(c) & \cmark & G-SceneCap & One-to-Many & 47.6 & 43.4 & 44.3 & 40.1 \\
(d) & \cmark & All data & One-to-One & 46.5 & 43.3 & 44.2 & \textbf{40.7} \\
(e) & \cmark & All data & One-to-Many  & \textbf{47.9} & \textbf{44.1} & \textbf{45.2} & 40.6 \\
\bottomrule
\end{tabular}
% }
\end{center}
\end{table*}

\begin{table*}[t]
\begin{center}
\caption{\textbf{Comparison of CLASP in phrase grounding tasks.} ``Point-supervision'' involves using point-level instance segmentation supervision. ``End-to-end models'' are those that can be trained either fully end-to-end or using separate detectors.} \label{tab: multi-tasks for scene-text alignment}
% \resizebox{\linewidth}{!}{
\begin{tabular}{lcccccccccc}
\toprule
\multirow{2}{*}{Method} & \multirow{2}{*}{ \shortstack{Point-\\supervision} } & \multirow{2}{*}{\shortstack{End-to\\-end}} & \multicolumn{2}{c}{ScanRefer Grd.} & \multicolumn{2}{c}{Multi3DRef Grd.} & \multicolumn{3}{c}{ScanNet200 Det.} \\ \cmidrule(lr){4-5} \cmidrule(lr){6-7} \cmidrule(lr){8-10}
 &  & & Acc@0.25 & Acc@0.5 & F1@0.25 & F1@0.5  & AP & AP@0.25 & AP@0.5 \\ \midrule
BUTD-DETR~\cite{butddetr} & \xmark & \xmark & 52.2 & 39.8 & -- & -- & -- & -- & --  \\
ViL3DRel~\cite{vil3drel} & \xmark & \cmark & 47.9 & 37.7 & -- & -- & -- & -- & -- \\
3D-VisTA~\cite{3dvista} & \xmark & \cmark & 50.6 & 45.8 & -- & -- & -- & -- & -- \\
M3DRef-CLIP~\cite{multi3drefer} & \xmark & \xmark & 51.9 & 44.7 & 42.8 & 38.4 & -- & -- & -- \\ 
Mask3D~\cite{schult2023mask3d} & \cmark & \cmark & -- & -- & -- & -- & 27.4 & 37.0 & \textbf{42.3} \\ \midrule
CLASP & \cmark & \cmark & 53.2 & 46.7 & 51.5 & 47.3 & 27.2 & 36.4 & 42.1  \\ 
% $2\times$ schedule & \cmark & \cmark & \textbf{54.3} & \textbf{47.4} & \textbf{54.6} & \textbf{49.6} & \textbf{27.5} & \textbf{37.0} & 41.4 \\
\bottomrule
\end{tabular}
% }
    
\end{center}
\end{table*}

\begin{table*}[t]
\caption{\textbf{Effects of pre-training datasets for CLASP.} $\dagger$ represents using ground-truth boxes.} \label{tab: effects of clasp}
\begin{center}
% \resizebox{1.\linewidth}{!}{
\begin{tabular}{ccclcccc}
\toprule
\multirow{2}{*}{\#} & \multirow{2}{*}{\shortstack{Pre-training\\Method}} & \multirow{2}{*}{\shortstack{Correspondence\\Level}} & \multirow{2}{*}{Training Data} & \multicolumn{2}{c}{ScanRefer Grd.} & \multicolumn{2}{c}{Multi3DRef Grd.} \\ \cmidrule(lr){5-6} \cmidrule(lr){7-8}
 & & & & Acc@0.25 & Acc@0.5 & F1@0.25 & F1@0.5 \\ \midrule
(a) & {\color[HTML]{969696}M3DRef-CLIP}$^\dagger$ & Sentence & Multi3DRef & \multicolumn{2}{c}{ {\color[HTML]{969696}55.0}} &  {\color[HTML]{969696}--} &  {\color[HTML]{969696}--} \\ \midrule
(b) & \multirow{2}{*}{3D-VisTA} & \multirow{2}{*}{Sentence} & ScanScribe & 41.1 & 36.4 & -- & -- \\ 
(c) & & & (b) w. ScanRefer & 50.6 & 45.8 & -- & -- \\ \midrule
(d) & \multirow{5}{*}{CLASP} & \multirow{5}{*}{Phrase} & G-SceneCap & 46.2 & 35.9 & 39.3 & 36.1 \\ 
(e) & & & (d) w. G-ScanRefer & 49.7 & 44.4 & 39.2 & 35.9 \\ 
(f) & & & (e) w. G-Detection & 51.1 & 44.8 & 40.2 & 37.0 \\
(g) & & & (f) w. G-Multi3DRef & \textbf{53.2} & \textbf{46.7} & \textbf{51.5} & \textbf{47.3} \\ 
(h) & & & (g) w/o. G-SceneCap & 51.1 & 45.4 & 48.5 & 43.1 \\
\bottomrule
\end{tabular}
% }
\end{center}
\end{table*}

\subsection{Main Results}

\noindent\textbf{Baselines.} Existing baseline methods for downstream tasks are categorized into discriminative and generative approaches in Table~\ref{tab: 3d llm performance}. 

\begin{itemize}[leftmargin=0.14in]
    \item  \textit{Discriminative models} focus on single-task outputs with task-specific heads, covering single-object~\cite{vil3drel, mvt, 3dvg-transformer} and multi-object grounding~\cite{multi3drefer}. 3D-VisTA~\cite{3dvista} pre-trains a unified 3D backbone followed by task-specific fine-tuning, while Mask3D~\cite{schult2023mask3d}, a SOTA transformer-based 3D detector, is trained with instance segmentation heads.
    \item  \textit{Generative models} include recent 3D LMMs~\cite{chat3d, chat3dv2, ll3da, 3dllm} \textit{without task heads}, where specialist models require task-specific fine-tuning, and generalist 3D LMMs~\cite{ll3da, embodiedgeneralist} are trained in unified language format.
\end{itemize}

% \todo{Discriminative models are designed for single-task outputs with structured formats (e.g., 2D detection outputs [x, y, h, w, category]). In contrast,generative models support open-ended outputs and can leverage multi-task datasets. Despite recent advances in 2D multi-modal models [2], aperformance gap remains.} 

\noindent\textbf{Result analysis.}
As shown in Table~\ref{tab: 3d llm performance}, Grounded 3D-LLM, \textit{without task-specific fine-tuning}, outperforms previous LLM-based models in most metrics, positioning it as a promising candidate for a unified framework in 3D scene understanding. Compared to prior 3D LMMs, it achieves the best grounding results, improving by over 10 points in both single and multiple object grounding. Additionally, it attains comparable outcomes in language-based tasks, even against LL3DA, which is limited to text output. Notably, it achieves leading performance with a CIDEr score of 75.9 and a BLEU-4 score of 13.2 in ScanQA. Existing generalist 3D LMMs, such as LEO~\cite{embodiedgeneralist} and LL3DA~\cite{ll3da}, support language tasks but are limited in localization. Furthermore, leveraging the one-to-many correspondence ability of the referent token, we report that Grounded 3D-LLM performs 3D instance segmentation, achieving 12.1 mAP. Despite its inferior performance compared to prior expert models like Mask3D, it demonstrates the potential to unify pure 3D vision tasks, with further improvements remaining as future work.

% \todo{why it performs worse than CLASP, Grounded 3D-LLM falls short of CLASP in localization tasks due to two main factors:
% 1.
% the unified modeling of multiple tasks has a trade-off between localization tasks and language tasks, and the randomly sampled diverse instructiontemplates (Section G) are used during training.
% 2.
% Grounded 3D-LLM (Tiny-Vicuna-1B) was fine-tuned using only a few LoRA layers, amounting to fewer than 200 MB of trainable parameters.}

\subsection{Ablation Studies}

\noindent\textbf{Diversity of task-specific instruction-following templates.}
Typical computer vision tasks such as language grounding, object detection, and dense captioning require instruction-following templates to better adapt to pre-trained weights of large language models. We assess the impact of diverse instruction-following templates on language tasks in Table \ref{tab: diverse templates}. Results show that diverse instruction-following templates consistently yield significant improvements in tasks like dense captioning and 3D QA. Notably, in template-based captioning tasks (Scan2Cap), both CIDEr@0.25 and CIDEr@0.5 see improvements exceeding $8$ points. Employing a single instruction-following template also leads to unstable results in repeated experiments if diverse templates are not used. Therefore, the diversity of instruction-following templates is crucial for sustaining language capabilities in visual tasks.

\noindent\textbf{Ablation studies of the \textit{referent} token for grounding.} 
As shown in Table~\ref{tab: effects of clasp for llm}, comparing line (a) with others indicates that directly adapting existing closed-set detectors is ineffective for integrating \textit{referent} tokens in Grounded 3D-LLM. Incorporating a grounded scene-text dataset for pre-training boosts performance to $43.2$ Acc@0.25 and $40.2$ F1@0.25. With only G-SceneCap for CLASP pre-training, Grounded 3D-LLM already achieves comparable results to all pre-training data.

Additionally, we analyze two types of \textit{referent} tokens: one-to-one (d), where each object is matched with a unique token via Hungarian matching, and one-to-many (e), where a single token represents multiple objects. The latter generally yields better results.

\noindent\textbf{Direct comparison of CLASP in phrase grounding tasks.} CLASP performs phrase-level scene-text alignment for pre-training, allowing for direct comparison with previous grounding and detection methods. As shown in Table~\ref{tab: multi-tasks for scene-text alignment}, CLASP excels in 3D grounding and detection benchmarks, particularly in ScanRefer grounding, Multi3DRef multi-object grounding, and ScanNet-200 detection, showcasing its enhanced phrase grounding capabilities compared to earlier models. Notably, in the Multi3DRef grounding task, CLASP outperforms specialized models by over 8 points, demonstrating its strong multi-object handling. 

% With a doubled training schedule, CLASP surpasses all competitors in grounding accuracy and matches the Mask3D model in detection performance. In this study, we primarily use the $1\times$ model for Grounded 3D-LLM, as the $2\times$ model, while better at grounding, underperforms in subsequent language tasks.

\noindent\textbf{Dataset ablations for CLASP.} Table~\ref{tab: effects of clasp} further ablates the effects of datasets for the scene-text alignment: 

\noindent\textit{(i) Zero-shot text evaluation.} Without training on ScanRefer or Multi3DRef language data, our CLASP model (d) proves effective compared to previous models (a) and (b). Notably, despite 3D-VisTA's data generation adhering to sentence-level grounding similar to ScanRefer, our model exceeds it by 4.9 points in Acc@0.25. Furthermore, in experiment (d), our direct evaluation on the Multi3DRef dataset demonstrates robust multi-object grounding ability. 

\noindent\textit{(ii) Multi-dataset evaluation.} We further investigate the effects of joint training with grounded scene-text datasets, progressively integrating G-ScanRefer (e), G-Detection (f), and G-Multi3DRef (g). This integration boosts performance, achieving 53.2 in Acc@0.25 on ScanRefer and 51.5 in F1@0.25 on Multi3DRef. Notably, Comparison between (g) and (h) shows that the G-SceneCap dataset consistently raises performance by +2.1 points in ScanRefer Acc@0.25 and +3.0 points in Multi3DRef F1@0.25, highlighting its effectiveness.

\begin{table}[t]
\caption{\textbf{Comparison of cross-modal interactors (CMI).}} \label{tab: cross interactor}
\begin{center}
% \resizebox{\linewidth}{!}{
\begin{tabular}{ccccccccccc}
\toprule 
\multirow{2}{*}{\shortstack{Cross-\\interactor types}} & \multicolumn{2}{c}{ScanRefer} & \multicolumn{2}{c}{Scan2Cap} \\ \cmidrule(lr){2-3} \cmidrule(lr){4-5}
 & Acc@0.25 & Acc@0.5 & B-4@0.5 & C@0.5 \\ \midrule 
One-way & 47.2 & 44.0 & 34.9 & 70.4 \\
Bi-directional. & \textbf{47.9} & \textbf{44.1} & \textbf{35.5} & \textbf{70.6} \\
\bottomrule
\end{tabular}
% }
\end{center}
\end{table}

\noindent\textbf{Comparison of cross-modal interactors.}
Table~\ref{tab: cross interactor} ablates two key multi-modal interaction components—cross-modal interactors of Fig.~\ref{fig: cmi design}: text-to-object one-way CMI and bidirectional CMI. The results indicate that incorporating bidirectional cross-attention modules in cross-modal interactors enhances grounding performance (with a 0.7 improvement in Acc@0.25 on ScanRefer and a 0.6 improvement in BLEU-4@0.5 on Scan2Cap) and dense captioning metrics. This improvement suggests that deeper interactions between modalities provide greater benefits for CLASP.

\section{Conclusion}\label{sec: conclusion}
We explore the potential of large multi-modal 3D models to unify various downstream 3D vision tasks in unified language modeling. By interpreting scene referents as special language tokens, Grounded 3D-LLM connects scene objects or regions with language, offering a natural way to localize noun phrases in the 3D environment. We demonstrate its general effectiveness across various 3D vision tasks; however, it shows limitations in performance compared to previous expert models, highlighting the disparity between specialist and generalist approaches. Scaling scene-text data to bridge this gap is an objective for future work.


\bibliographystyle{IEEEtran}
\bibliography{egbib}

% \input{bio/bio}

% \appendix
\captionsetup[table]{labelformat={default},labelsep=period,name={Table A -}}
\captionsetup[figure]{labelformat={default},labelsep=period,name={Figure A -}}

\renewcommand{\thesubsection}{\Alph{subsection}}

\onecolumn
\begin{center}\Large\bf
Supplementary Files for Grounded 3D-LLM
\end{center}

% \maketitle

The supplementary materials are organized as follows:
\begin{enumerate}[leftmargin=*]
\item Additional ablation studies are included in Sec.~\ref{sec: additional ablation studies}.
\item Hyper-parameters of training are included in Sec.~\ref{sec:more implementation-details}.
\item Visualizations of results and failure cases for downstream tasks are presented in Section~\ref{sec: result visualization}.
\item Details on grounded scene-text data generation, visualizations of dataset statistics, and data generation prompts, are illustrated in Sec.~\ref{sec: grounded scene caption stat}.
\item Multi-task instruction-following templates are described in Sec.~\ref{sec: llm templates}.
\item Extension to embodied dialogue and planning data is shown in Sec.~\ref{sec: embodied dialog planning}
\end{enumerate}

\section{Additional Ablation Studies} \label{sec: additional ablation studies}

\noindent\textbf{Effects of LLM model sizes.} 
As shown in Table A-\ref{tab: model size}, we compare the performance of GPT-2, Tiny-Vicuna-1B, Vicuna-7B, and Vicuna-13B. Smaller models like GPT-2 struggle with soft \textit{referent} token learning. In contrast, Grounded 3D-LLM models, ranging from 1B to 13B parameters, significantly outperform GPT-2 and show marginal improvements on Scan2Cap, though they experience a slight decline on ScanQA. Notably, even with only the projection layer trainable (without LoRA), Grounded 3D-LLM (Vicuna-7B) achieves comparable performance, suggesting that the input scene embeddings are semantically rich and can be extracted by the LLM without intermediate trainable layers. Furthermore, the lack of improvement in language tasks with larger models indicates the current task scope may not fully challenge larger-scale LLMs.

\begin{table*}[h]
\caption{\textbf{Comparison of different LLMs.} For GPT-2, we train the overall backbones with a smaller language head. In models without LoRA, only the projection layers are trained.} \label{tab: model size}
\begin{center}
% \resizebox{\linewidth}{!}{
\begin{tabular}{lcccccccccc}
\toprule 
\multirow{2}{*}{LLM} & \multirow{2}{*}{LoRA} & \multicolumn{2}{c}{ScanQA}  & \multicolumn{4}{c}{Scan2Cap}& \multirow{2}{*}{\shortstack{Trainable\\Params}} \\ \cmidrule(lr){3-6} \cmidrule(lr){7-8}
 & & B-4 & C & B-4@0.25 & C@0.25 & B-4@0.5 & C@0.5 \\ \midrule 
GPT-2 & \cmark & 8.42 & 44.6 & 28.4 & 45.2 & 26.7 & 41.9 &  169 MB\\
Tiny-Vicuna-1B & \cmark & \textbf{13.2} & \textbf{75.9} & 36.9 & 74.1 & 35.0 & 70.2 & 157 MB \\
Vicuna-7B & \cmark & \textbf{13.3} & 74.1 & 37.8 & 74.3 & 35.4 & \textbf{70.8} & 364 MB \\
Vicuna-7B & \xmark & 12.4 & 70.9 & 37.4 & 74.2 & 34.7 & 69.9 & 330 MB \\
Vicuna-13B & \cmark & 12.9 & 73.2 & \textbf{38.5} & \textbf{74.6} & \textbf{35.9} & 70.4 & 487 MB \\
\bottomrule
\end{tabular}
% }
\end{center}
\end{table*}

\section{Implementation Details} \label{sec:more implementation-details}
The training hyper-parameters are detailed in Table A-\ref{tab: hyperparameters}. All models are trained on 8 NVIDIA A100 (80G) GPUs. 

\begin{table}[h] 
\caption{\textbf{Training hyper-parameters.} The instruction fine-tuning batch size comprises 5 scenes, each containing up to 200 instructions.}
\begin{center}
% \resizebox{\linewidth}{!}{
\begin{tabular}{cccccc}
\toprule 
Configuration & CLASP pre-training & Instruction fine-tuning \\ \midrule
Queries & 150 & 100 \\
Epochs & 600 & 50 \\ 
Batch size & 5 & 5$\times$200 \\
Learning rate & 1e-4 & 8e-4 \\
LR schedule & Cosine & Cosine \\
Optimizer & AdamW & AdamW  \\
\bottomrule
\end{tabular}
% } 
\end{center}
\label{tab: hyperparameters}
\end{table}

\section{Result Visualizations}
\label{sec: result visualization}

\begin{figure}[h]
  \centering
  \includegraphics[width=.7\linewidth]{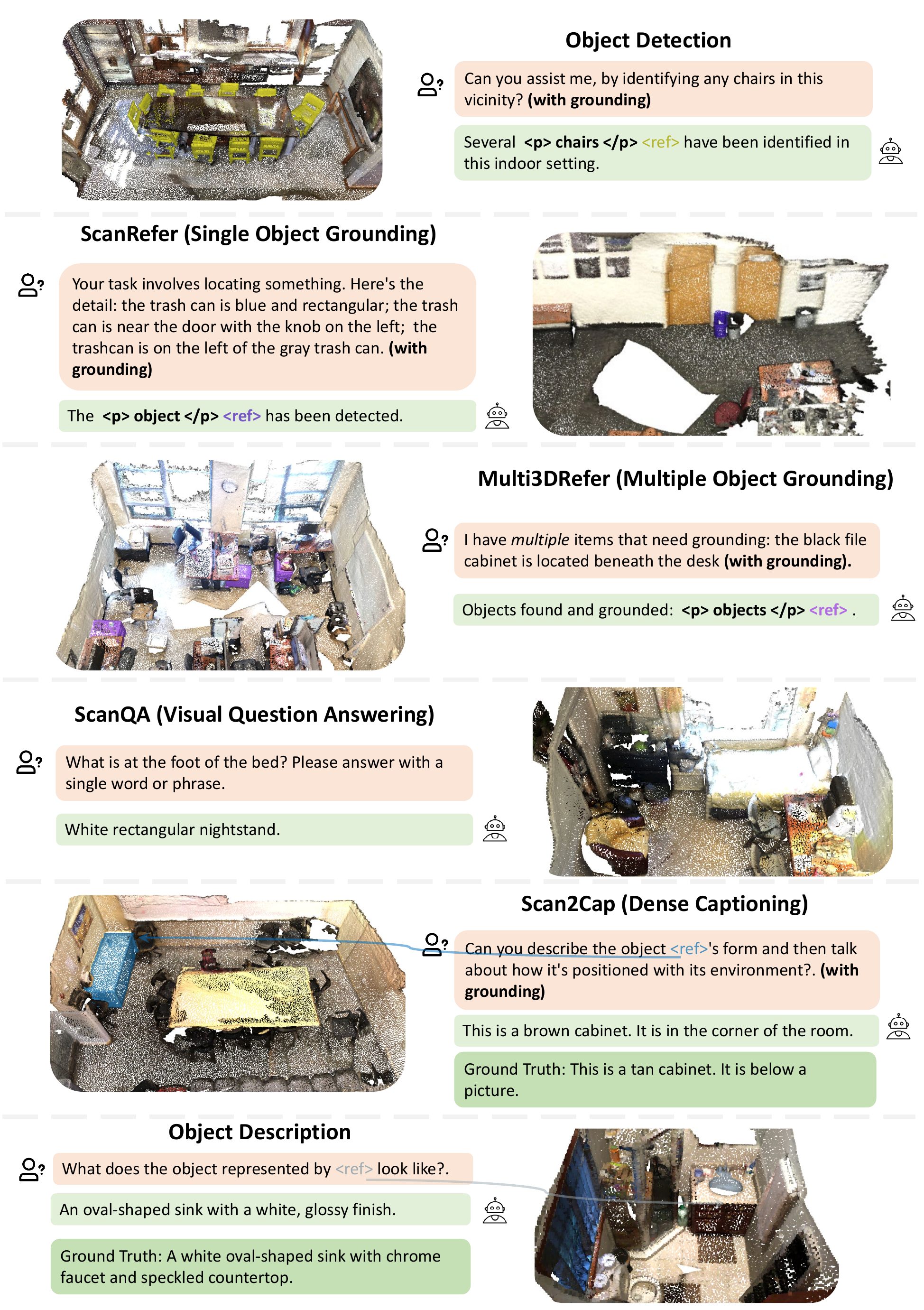}\vspace{-0.1in}
  \caption{\textbf{One model to solve all tasks.} Result visualization in multiple downstream benchmarks for Grounded 3D-LLM. Please zoom in for better visualization.}
  \label{fig: result visualization}
\end{figure}

\noindent\textbf{Result visualization.} We visualize the results in Fig. A-\ref{fig: result visualization} for a single model Grounded 3D-LLM that solves all downstream tasks, as opposed to per-task fine-tuning on downstream tasks\cite{3dvista, 3dllm, ll3da}. 

\begin{figure}[h]
  \centering
  \includegraphics[width=.7\linewidth]{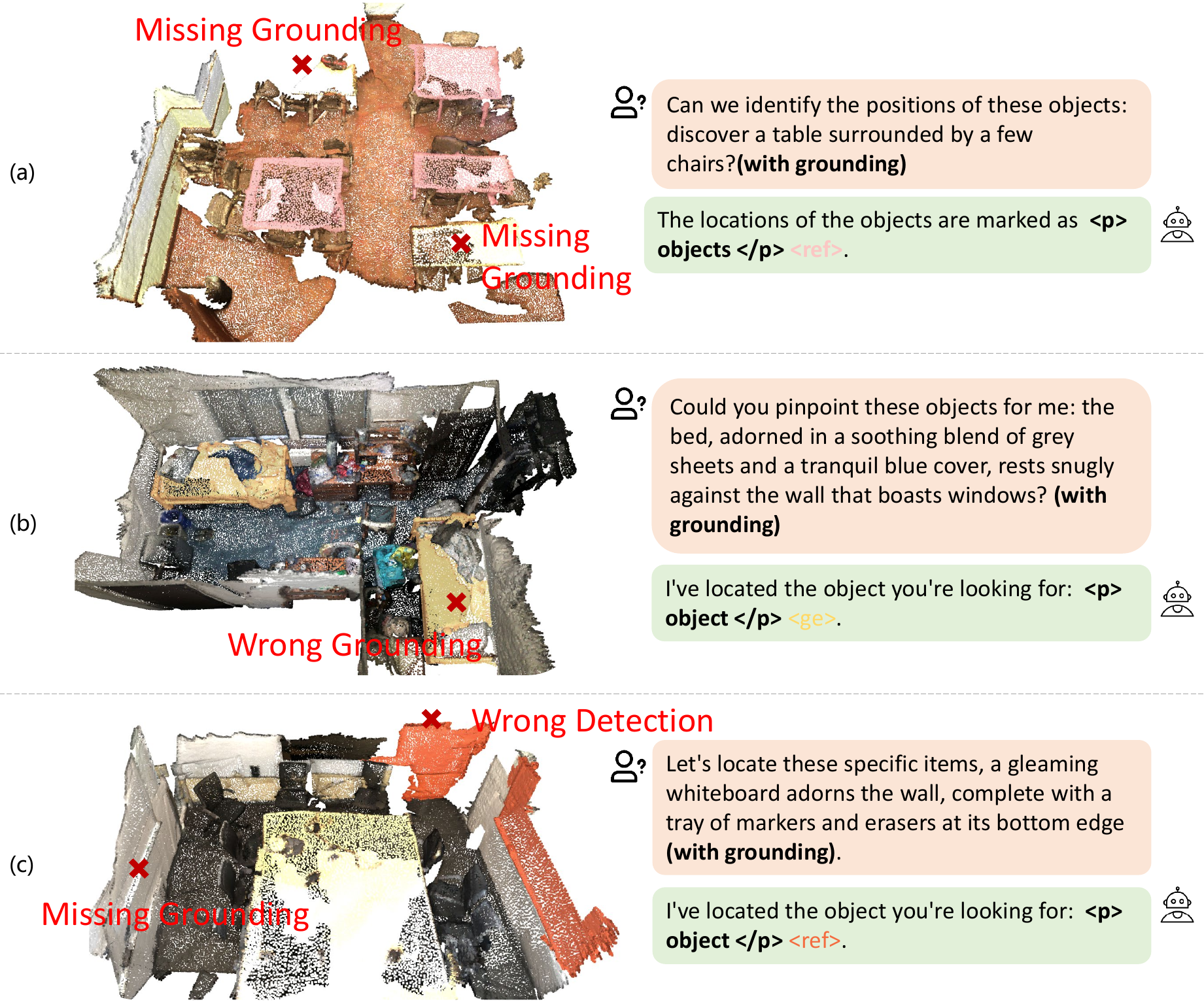}
  \caption{\textbf{Failure cases.} Please zoom in for better visualization.}
  \label{fig: failure case}
\end{figure}

\noindent\textbf{Failure cases.} As depicted in Fig. A-\ref{fig: failure case}, the primary reasons for our model's failures in the grounding task include missing grounding objects (a, c), excessive grounding of semantically incorrect objects (b), and errors in detection/segmentation (c).

\newpage
\section{Grounded Scene Caption Dataset} \label{sec: grounded scene caption stat}

\noindent\textbf{Data generation details.}
For the creation of Grounded Scene Captions (G-SceneCap), we selected the ScanNet-200 dataset due to its extensive range of semantic object categories. The data generation process employs heuristics to ensure that each local scene caption is concise, with fewer than 256 words. Initially, we identify each anchor object within a 2-meter radius to tally the objects. If this count exceeds 15, we reduce the radius to maintain this limit. Subsequent object selection is conducted through random sampling, with selection probabilities ranging from 0.6 to 0.9. This method prevents the selection of repetitive object types while preserving the anchor. Finally, for each caption, we generate program-derived relations that use existing objects as the anchor following SR3D~\cite{referit3d}. These relations are then integrated into the caption. The overall process always keeps both the anchor and target object IDs.

\noindent\textbf{Post processing.} As ChatGPT/GPT-4 may generate errors such as incorrect output formats or erroneous object IDs, we filter out these wrong captions from the dataset.

\begin{figure}[h]
  \centering
  \includegraphics[width=.8\linewidth]{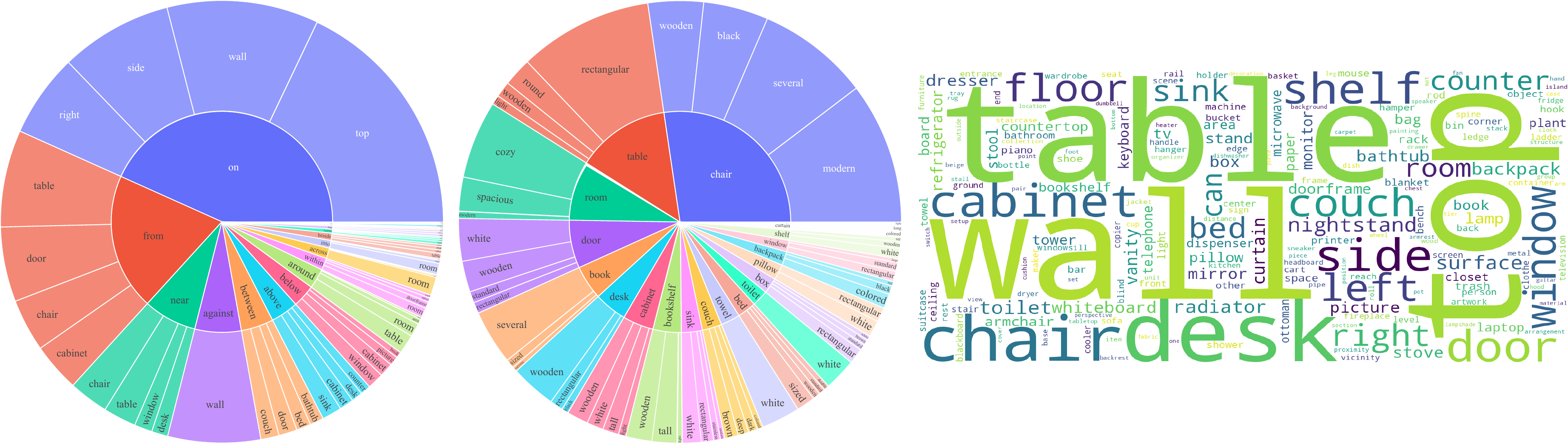}
  \caption{\textbf{Data statistics of grounded scene caption.} \textit{Left}: Diagram of spatial preposition-noun statistics, \textit{Middle}: Noun-adjective statistics diagram, \textit{Right}: Noun word cloud.} \label{fig: statistic 1}
\end{figure}

\noindent\textbf{Dataset language components visualization.} For the grounded scene caption dataset, we visualize in Fig. A-\ref{fig: statistic 1} the relationships between noun-adjective pairs and spatial preposition-noun pairs across all languages, along with a word cloud of all nouns, to demonstrate the diversity of our curated dataset.

\noindent\textbf{Dataset visualization.} 
We provide additional visualizations of grounded scene caption examples in Fig. A-\ref{fig: dataset visualization}. 

\begin{figure*}[t]
  \centering
  \includegraphics[width=0.95\linewidth]{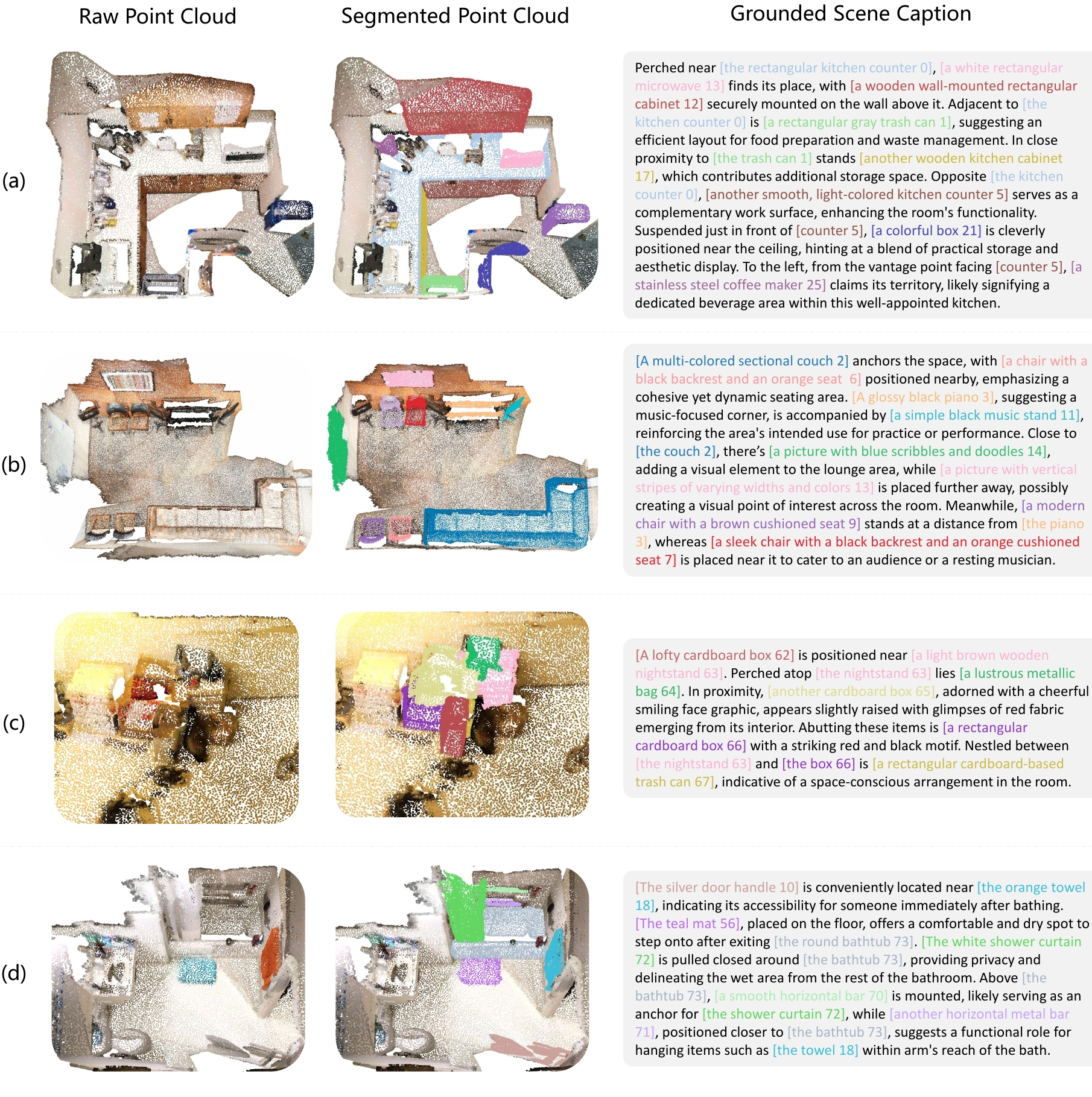}
  \caption{\textbf{Example visualization of grounded scene caption dataset.} The ``[ ]'' symbol marks each grounded phrase. The color of the phrase corresponds to the segmented point cloud.}
  \label{fig: dataset visualization}
\end{figure*}

\begin{figure}[htbp]
  \centering
  \includegraphics[width=0.7\linewidth]{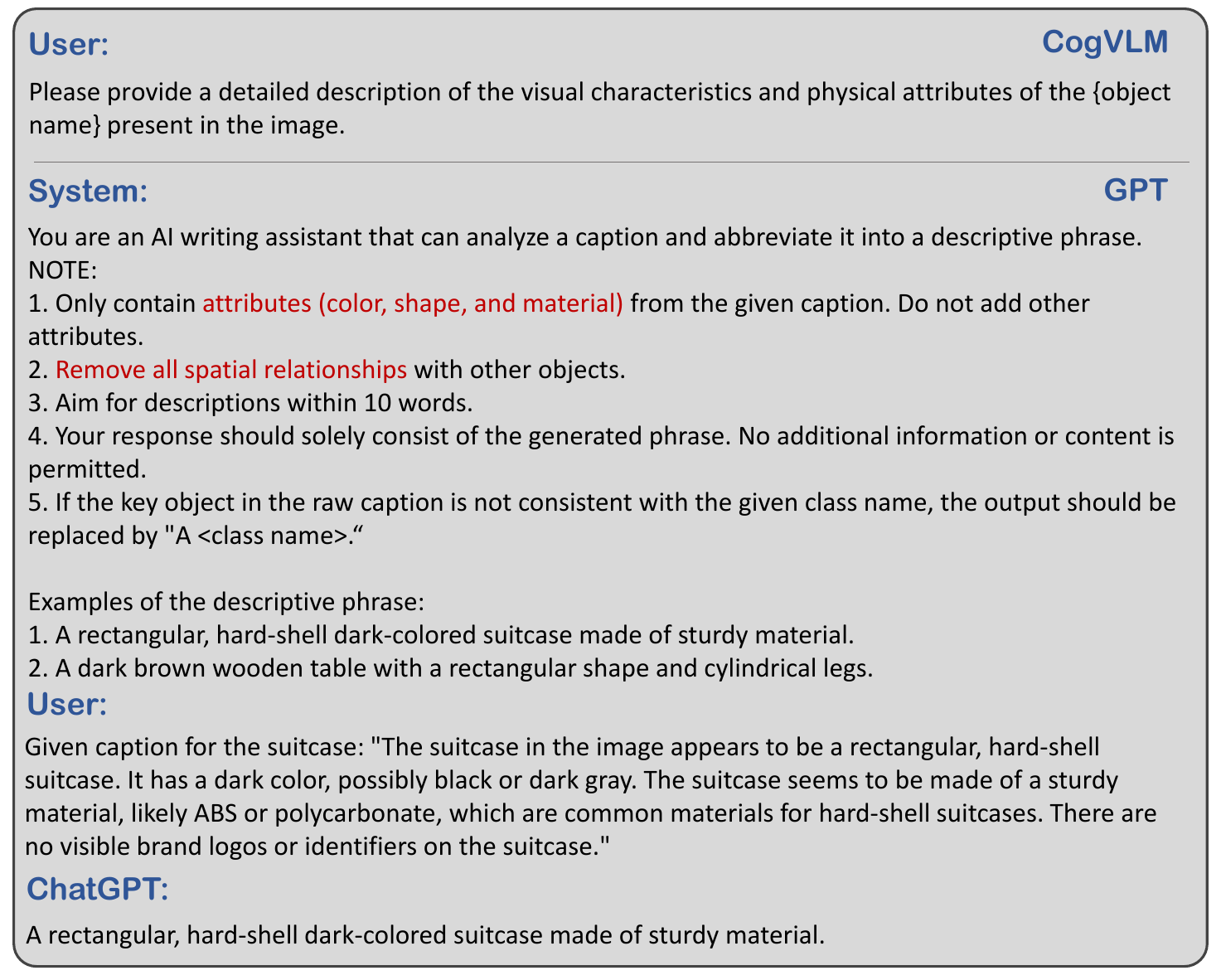}
  \caption{\textbf{Object caption prompt for VLM and ChatGPT (used in Step-1 of data generation).} The above is the prompt for CogVLM that utilizes the object labels from ScanNet-200. Below is the prompt for ChatGPT input, where ``System'' denotes the system prompt and ``User'' denotes user input.}
  \label{fig: object caption}
\end{figure}

\begin{figure}[htbp]
  \centering
  \includegraphics[width=0.7\linewidth]{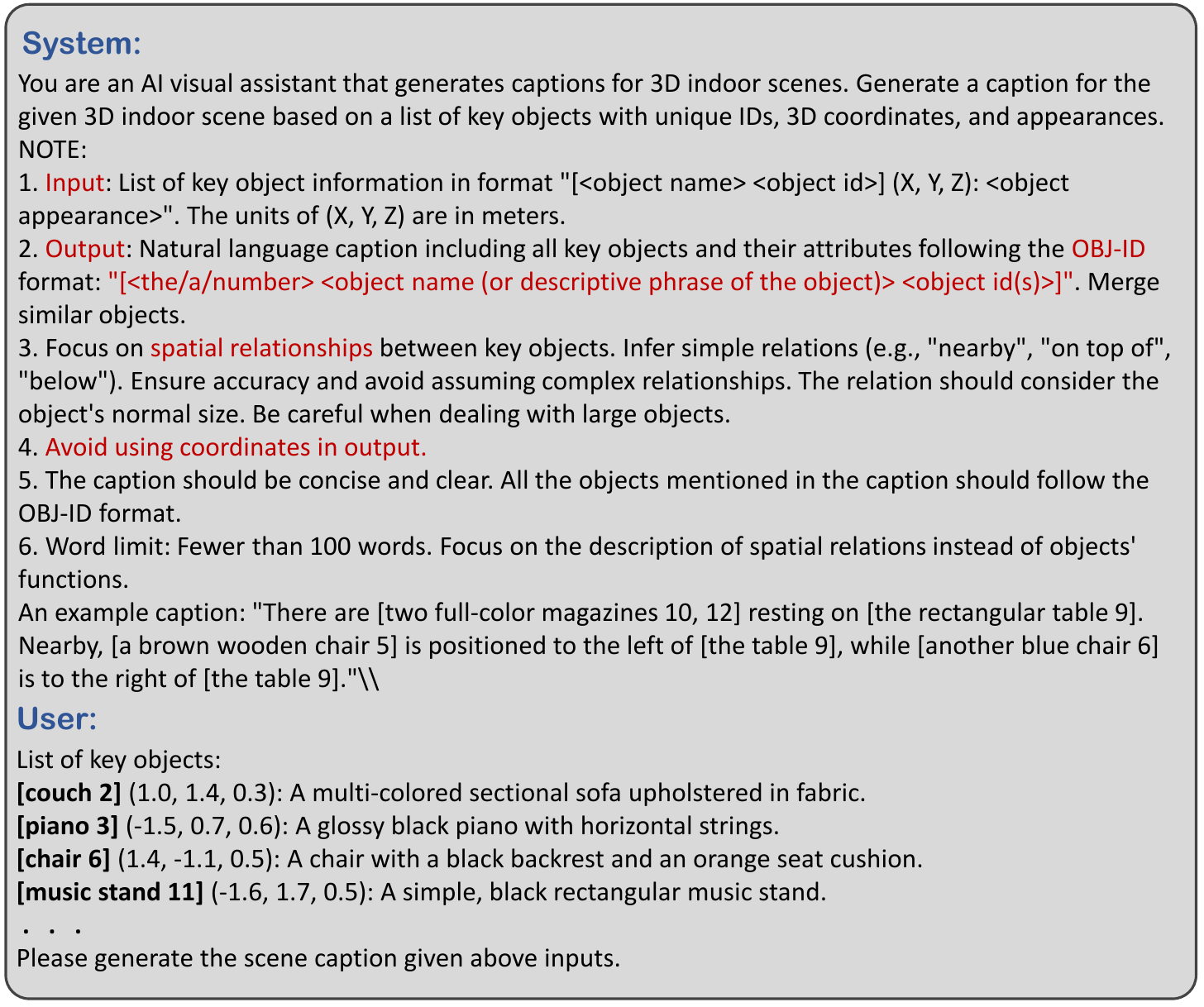}
  \caption{\textbf{GPT-4 prompt for scene caption (used in Step 2 of data generation).} ``System'' denotes the system prompt and ``User'' denotes user input for GPT-4. }
  \label{fig: scene caption}
\end{figure}

\begin{figure}[htbp]
  \centering
  \includegraphics[width=0.7\linewidth]{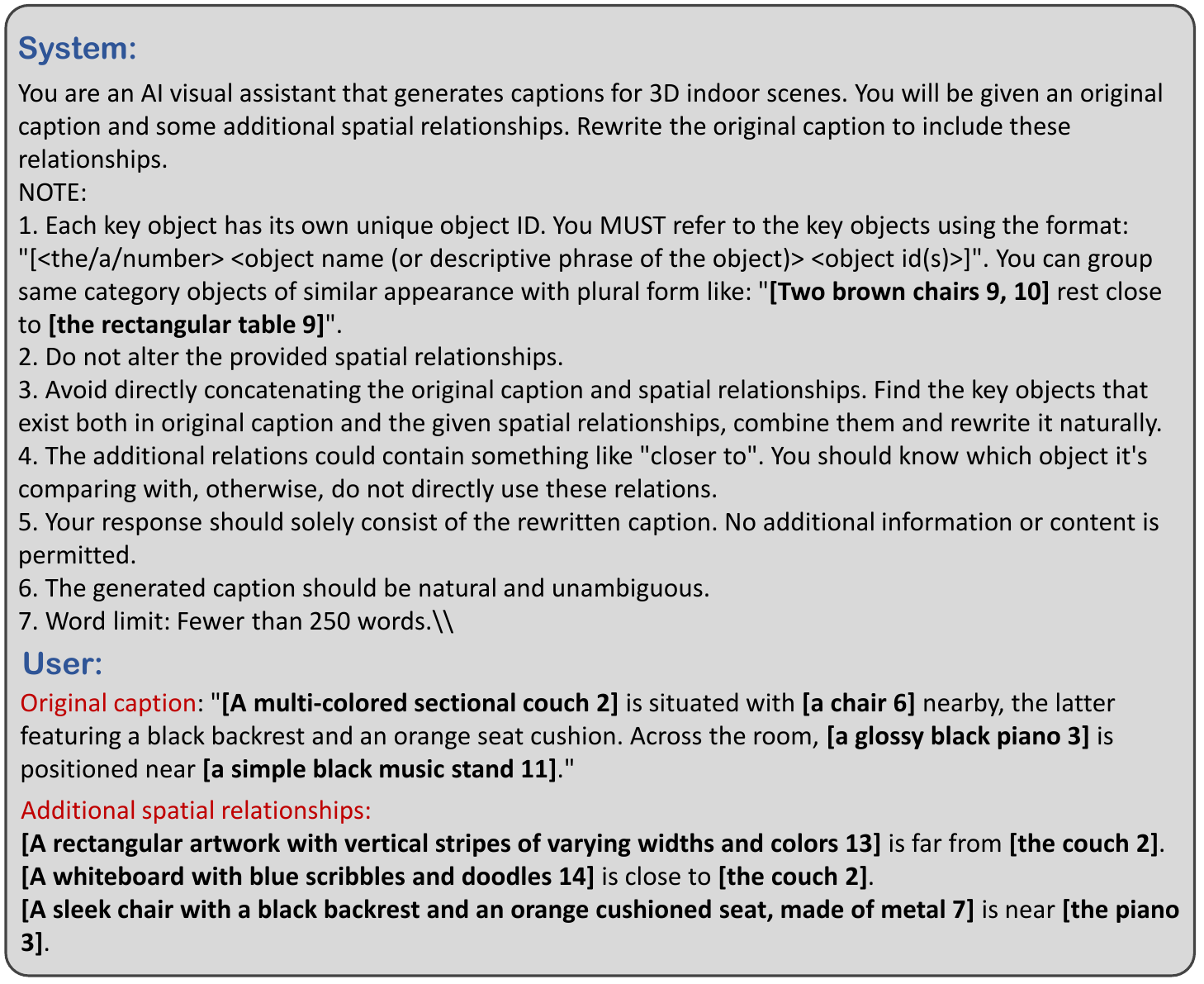}
  \caption{\textbf{GPT-4 prompt of including rule-based relation (used in Step 3 of data generation).} ``System'' denotes the system prompt and ``User'' denotes user input for GPT-4. }
  \label{fig: insert sr3d caption}
\end{figure}

\newpage
\noindent\textbf{Data generation prompts} Fig. A-\ref{fig: object caption} displays the object caption prompts, including the VLM captioning prompt~\cite{cogvlm} and the subsequent condensation of the caption into a descriptive phrase using ChatGPT. These object captions are then compiled to create local scene captions, as shown in Fig. A-\ref{fig: scene caption}, while retaining their object IDs. Finally, spatial relationships derived from program-generated code are incorporated using the prompt depicted in Fig. A-\ref{fig: insert sr3d caption}.

\section{Task-specific Instruction-following Templates}\label{sec: llm templates}

Grounded 3D-LLM simultaneously undertakes vision tasks—such as 3D visual grounding, 3D object detection—and language tasks, like question answering. Each task demands specific, diverse templates to enable the LLM to \textit{\textbf{naturally}} provide scene-aware answers tailored to particular dataset specifications. We show several instruction-following templates in the following figures.

\begin{figure}[htbp]
  \centering
  \includegraphics[width=.7\linewidth]{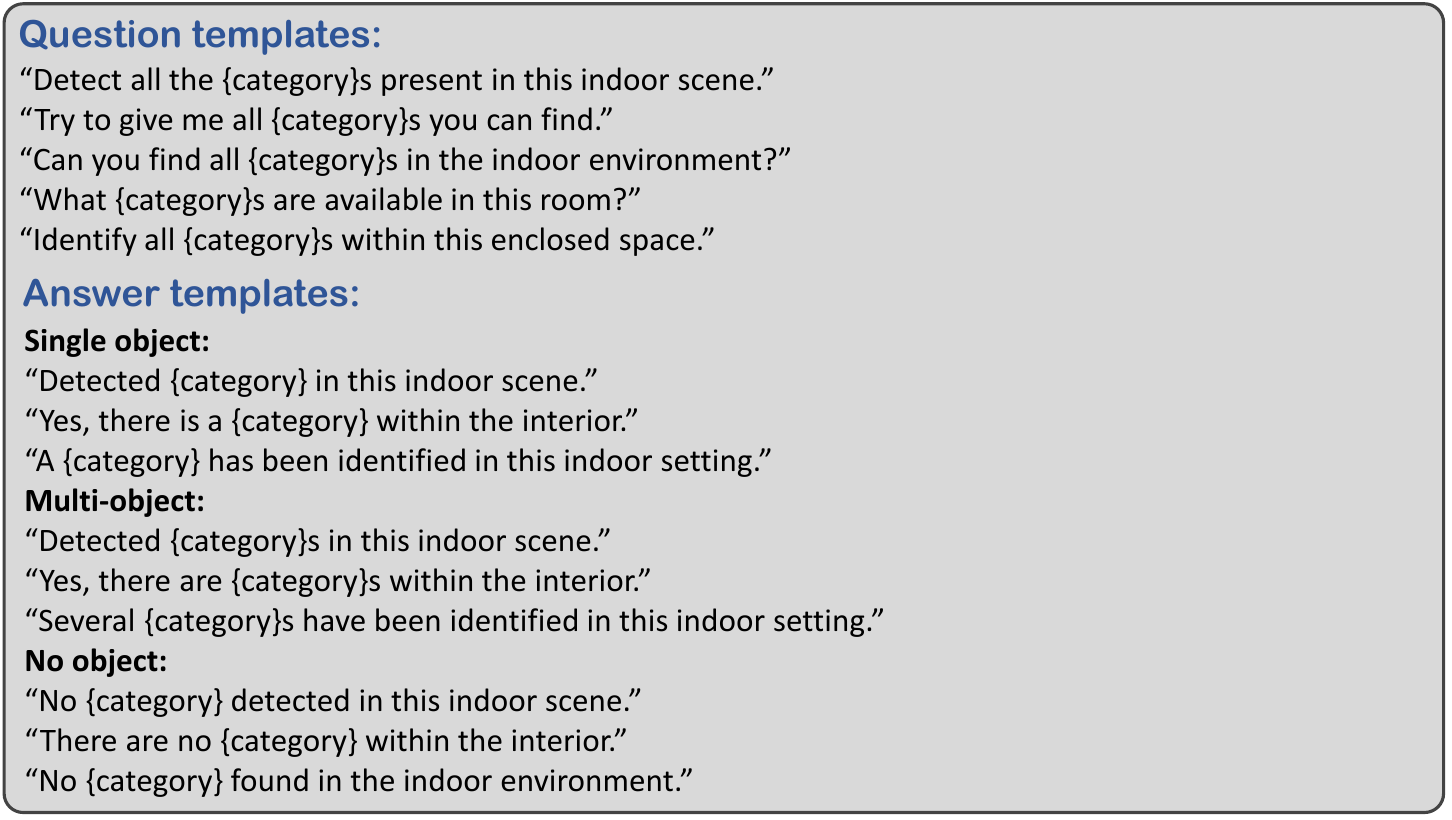}\vspace{-0.1in}
  \caption{\textbf{Instruction-following templates for object detection.} }
  \label{fig: detection template}
\end{figure}

\noindent\textbf{Object detection}. The detection tasks employ various question templates for each category, along with three answer templates to accommodate various detection outcomes -- single, multiple, or no objects detected -- as illustrated in Fig. A-\ref{fig: detection template}.

\begin{figure}[htbp]
  \centering
  \includegraphics[width=.7\linewidth]{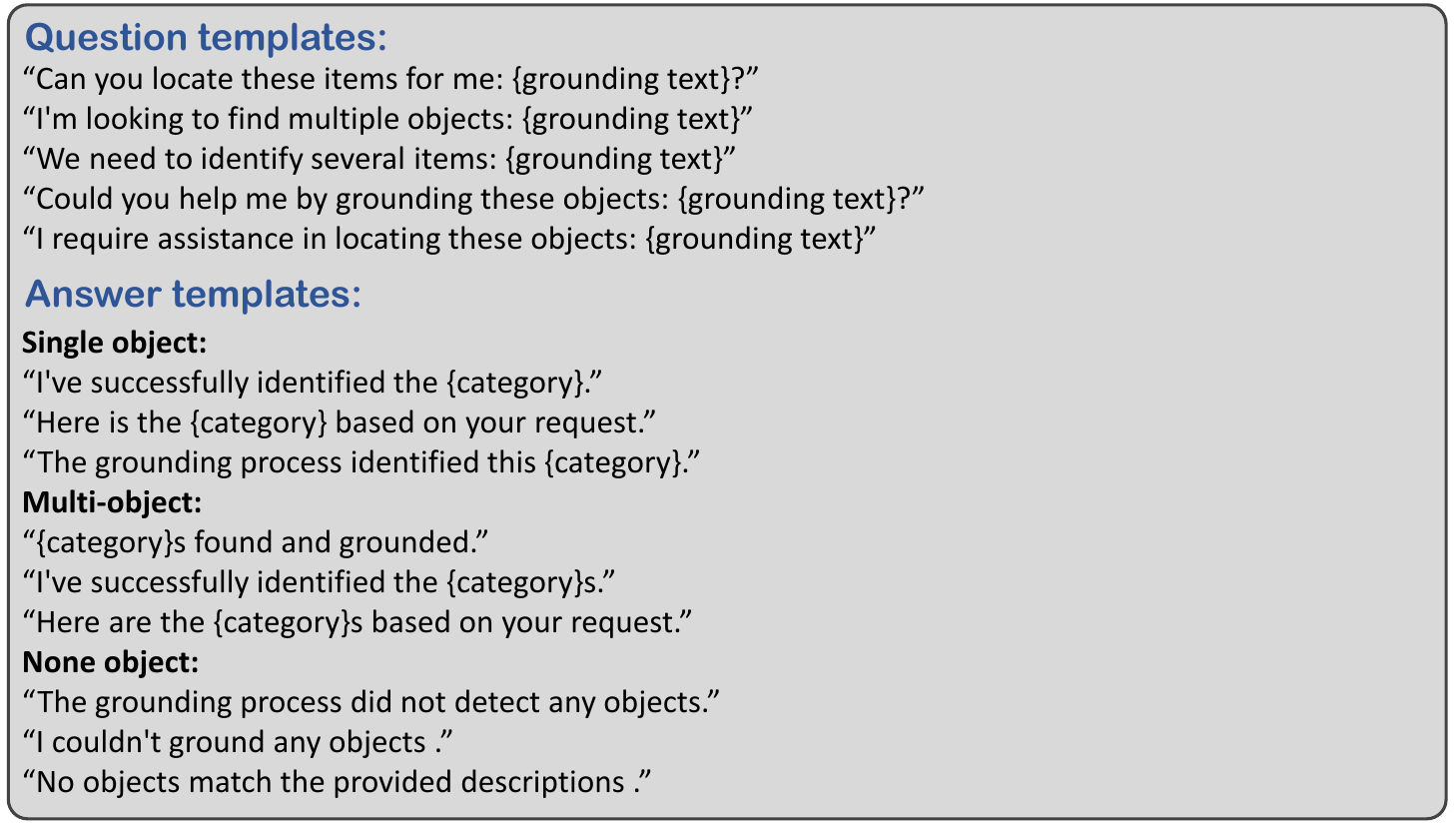}\vspace{-0.1in}
  \caption{\textbf{Instruction-following templates for single- or multi- object grounding.} }
  \label{fig: grounding template}
\end{figure}

\noindent\textbf{Single and multi-object grounding.} The language grounding's question template, as depicted in Fig. A-\ref{fig: grounding template}, directs the model to identify referred objects. For the ScanRefer grounding dataset, the answer templates indicate single-object grounding. The answer templates for the Multi3DRef dataset accommodate various grounding scenarios—single, multiple, or no grounded objects.

\begin{figure}[htbp]
  \centering
  \includegraphics[width=.7\linewidth]{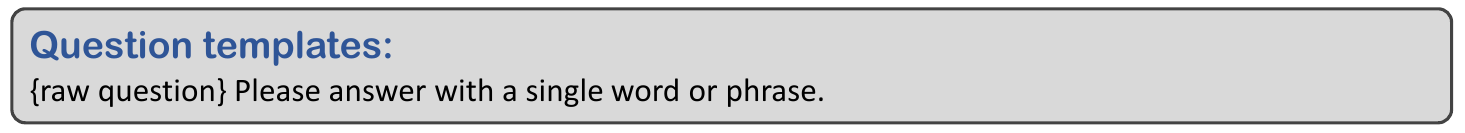}\vspace{-0.1in}
  \caption{\textbf{Instruction-following templates for ScanQA.} }
  \label{fig: scanqa template}
\end{figure}

\noindent\textbf{3D Question answering.} For ScanQA, we append suffixes to brief outputs (Fig. A-\ref{fig: scanqa template}) to indicate the outputs with only phrases or word output as the dataset annotation guideline.

\begin{figure}[htbp]
  \centering
  \includegraphics[width=.7\linewidth]{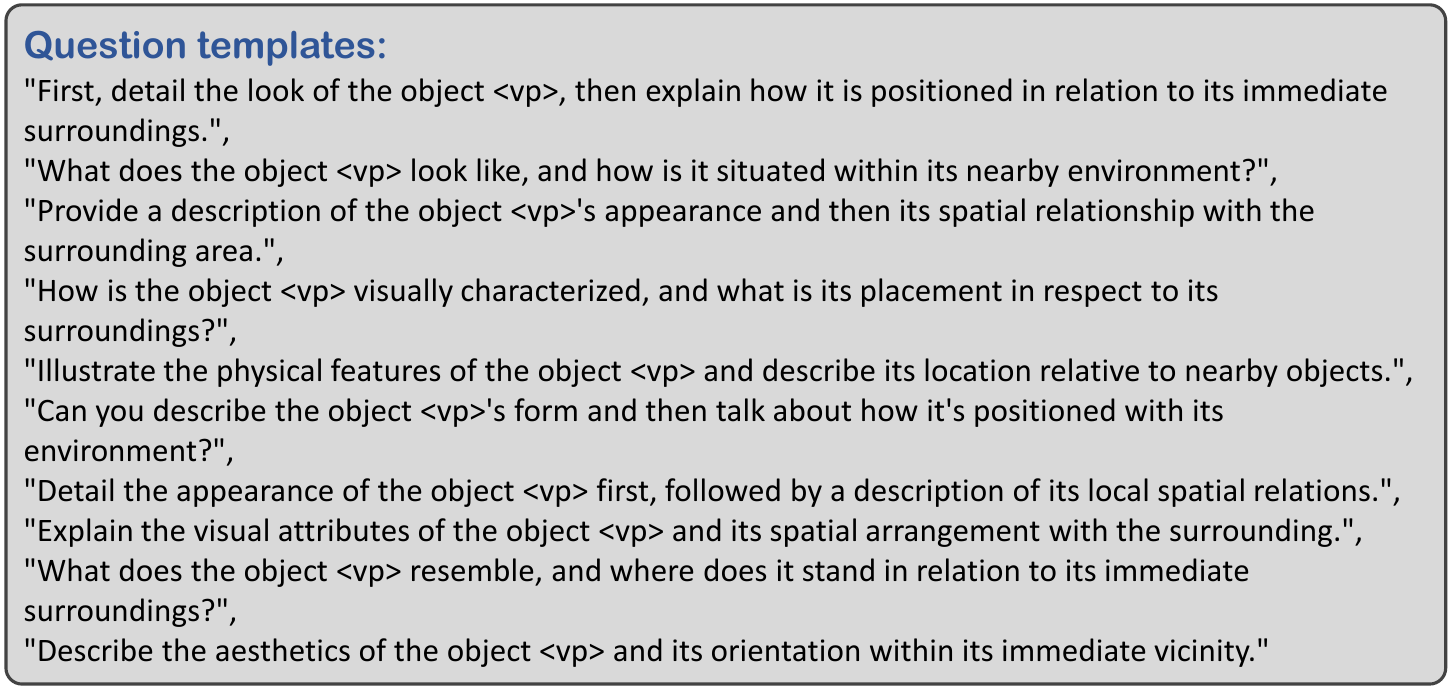}
  \caption{\textbf{Instruction-following templates for Scan2Cap.} } \vspace{-0.1in}
  \label{fig: scan2cap template}
\end{figure}

\noindent\textbf{Dense captioning.} Question templates for Scan2Cap require the model to describe object appearance followed by spatial relations, as illustrated in Fig. A-\ref{fig: scan2cap template}.

\section{Extension to embodied dialogue and embodied planning}\label{sec: embodied dialog planning}

\begin{figure}[htbp]
  \centering
  \includegraphics[width=0.7\linewidth]{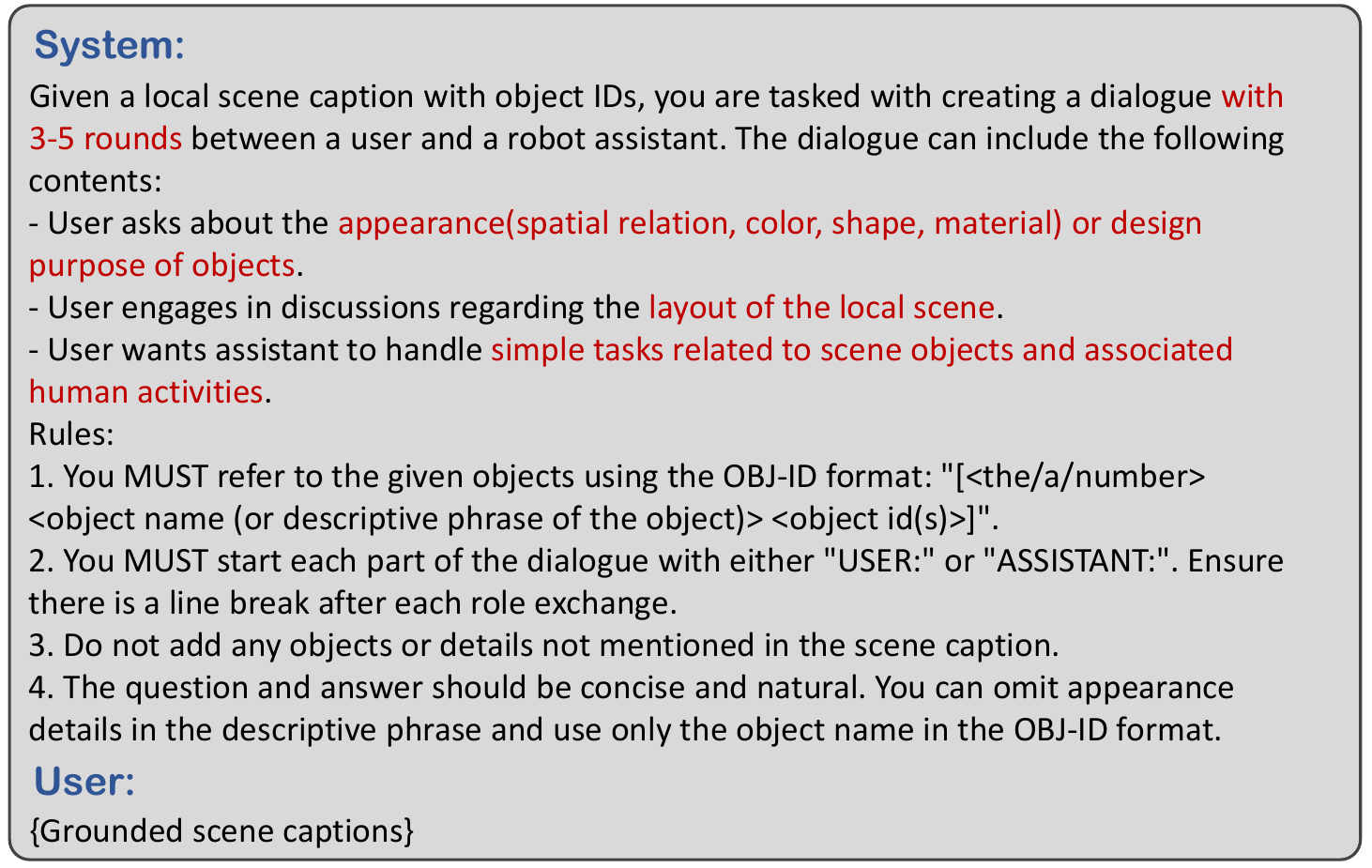}
  \caption{\textbf{GPT-4 Prompt of embodied dialogue.} ``System'' denotes the system prompt and ``User'' denotes user input for GPT-4. ``\{Grounded scene caption\}'' is filled with our grounded scene caption data. }
  \label{fig: dialog prompt}
\end{figure}

\begin{figure}[htbp]
  \centering
  \includegraphics[width=0.7\linewidth]{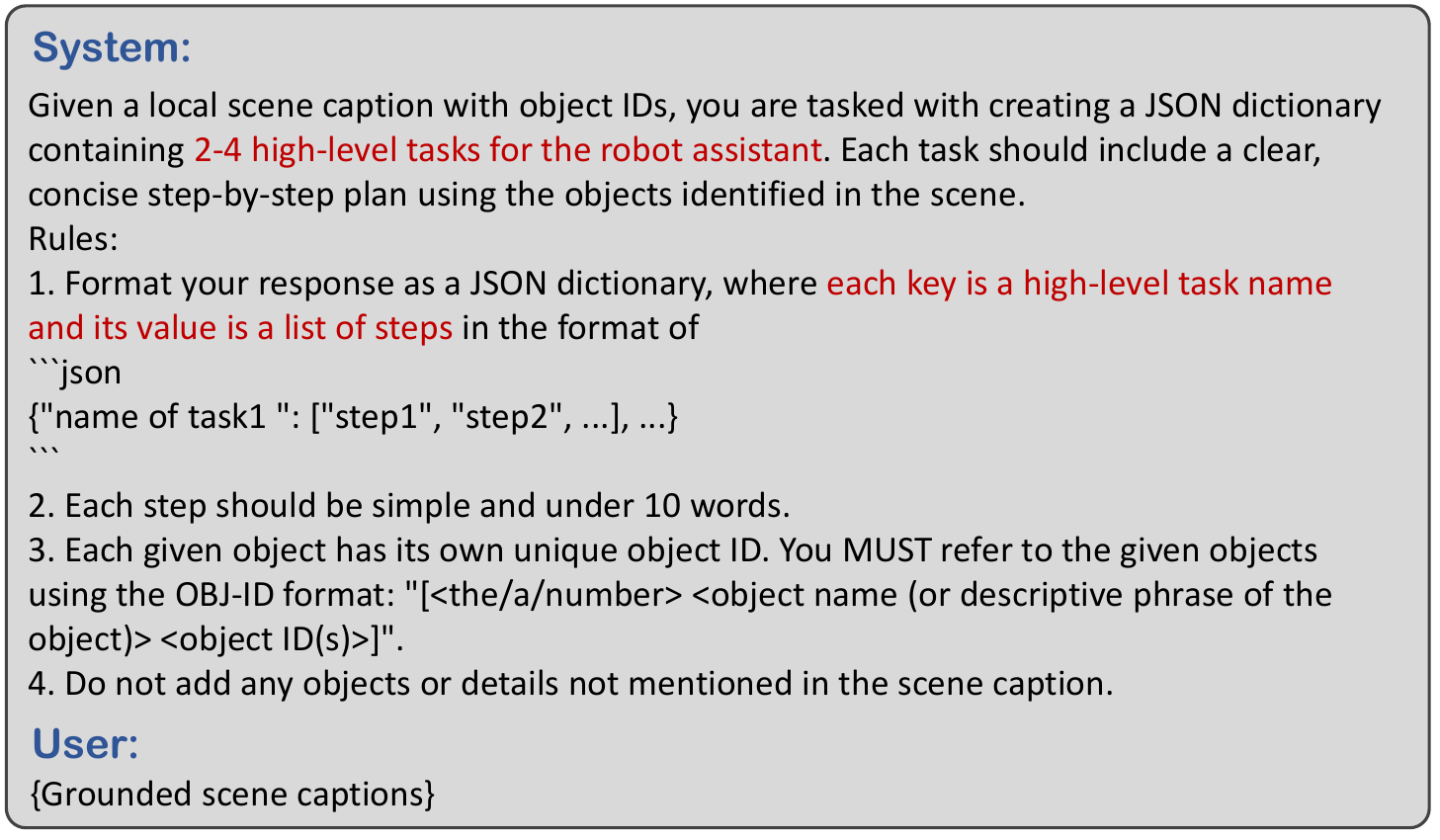}
  \caption{\textbf{GPT-4 Prompt of embodied planning.} ``System'' denotes the system prompt and ``User'' denotes user input for GPT-4. ``\{Grounded scene caption\}'' is filled with our grounded scene caption data. }
  \label{fig: planning prompt}
\end{figure}

Fig~A-\ref{fig: dialog prompt} and A-\ref{fig: planning prompt} list the prompts for grounded scene caption-based embodied dialogue and planning. The training dataset includes 11.9K dialogue examples and 4.4K planning examples. 

\begin{figure}[h]
  \centering
  \includegraphics[width=.7\linewidth]{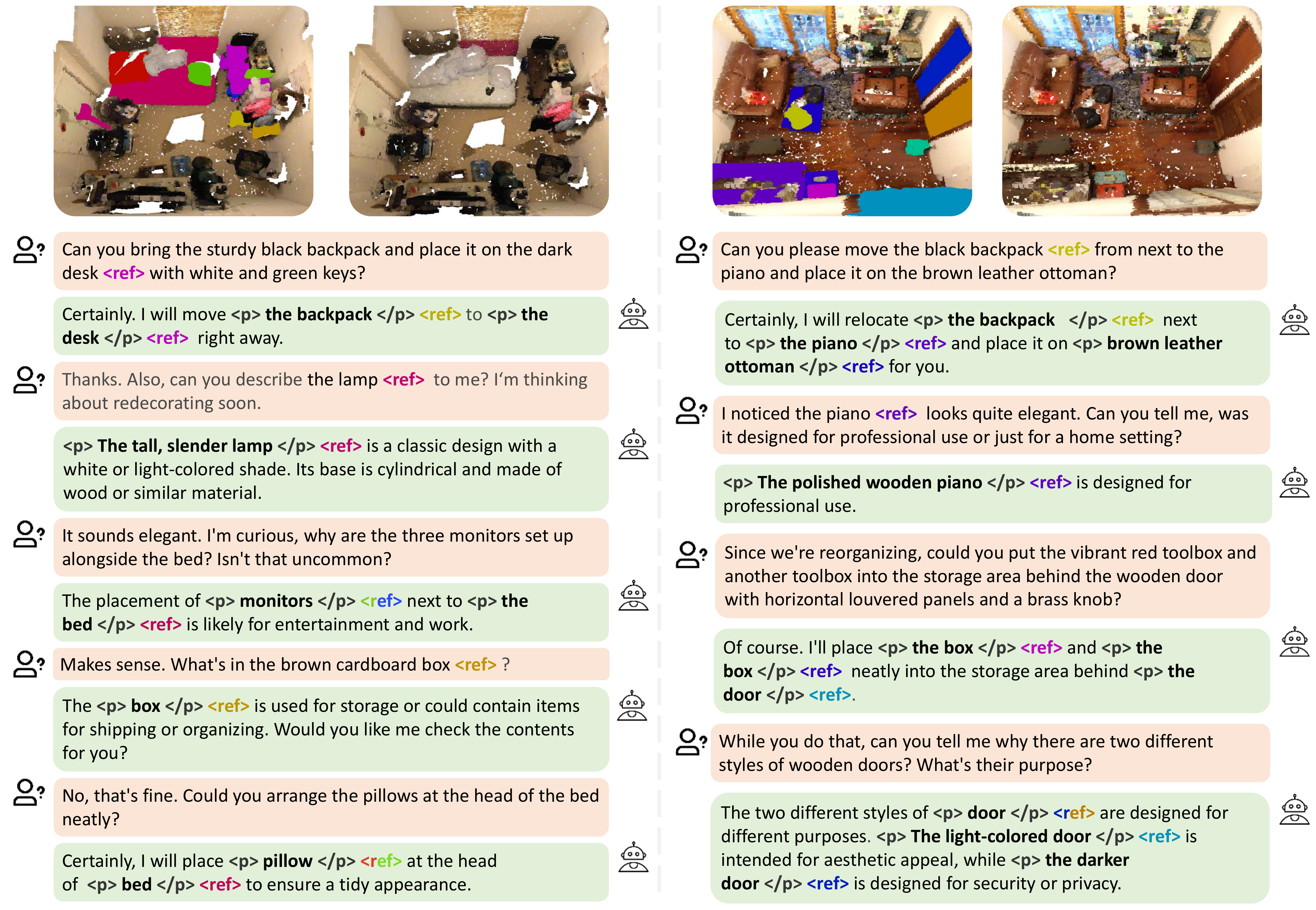}
  \caption{\textbf{Result visualization for embodied dialogue.} Please zoom in for better visualization.} \vspace{-0.1in}
  \label{fig: embodied dialog result}
\end{figure}

\begin{figure}[h]
  \centering
  \includegraphics[width=.7\linewidth]{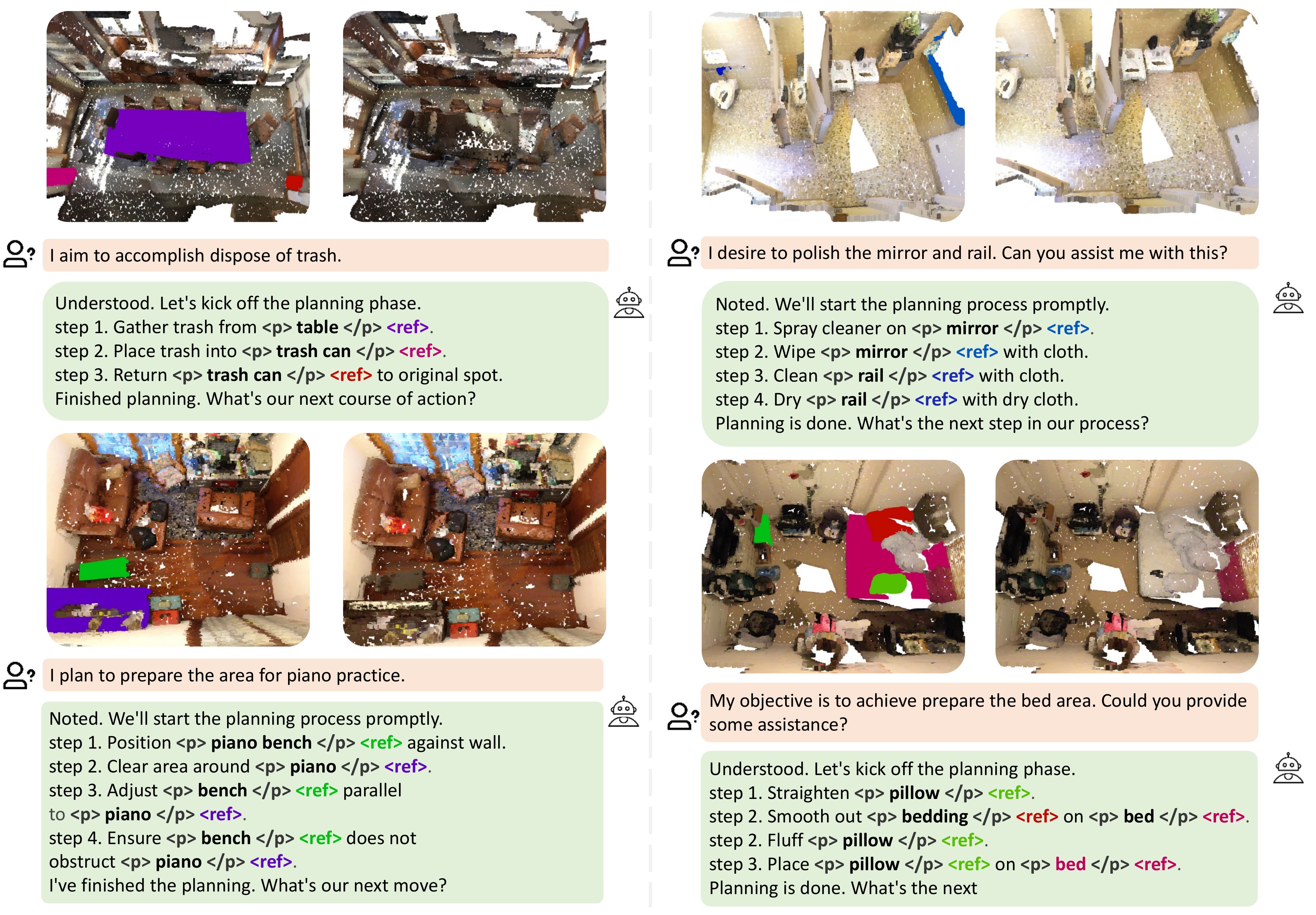}
  \caption{\textbf{Result visualization for embodied planning.} Please zoom in for better visualization.} \vspace{-0.1in}
  \label{fig: embodied plan result}
\end{figure}

\noindent\textbf{Result visualization.} As there is no well-defined benchmark for evaluating both phrase grounding and language understanding, we directly visualize the results of the jointly trained Grounded 3D-LLM for embodied dialogue (Fig.~\ref{fig: embodied dialog result}) and embodied planning (Fig.~\ref{fig: embodied plan result}). These examples demonstrate the potential of Grounded 3D-LLM's grounding capability to enhance robot navigation and manipulation.

\end{document}